\documentclass{article} % For LaTeX2e
\usepackage[final]{colm2026_conference}

\usepackage{microtype}
\usepackage{hyperref}
\usepackage{url}
\usepackage{booktabs}
\usepackage{graphicx}
\usepackage[most]{tcolorbox}
\usepackage{listings}
\usepackage{xcolor}
\usepackage{makecell}
\usepackage{pifont}
\usepackage{booktabs}

\lstdefinestyle{promptstyle}{
  basicstyle=\ttfamily\small,
  breaklines=true,
  columns=fullflexible,
  keepspaces=true,
  showstringspaces=false,
  frame=none
}

\newtcblisting{promptbox}[2][]{
  breakable,
  enhanced,
  colback=white,
  colframe=black,
  boxrule=0.8pt,
  arc=2mm,
  left=2mm,
  right=2mm,
  top=1mm,
  bottom=1mm,
  listing only,
  listing options={style=promptstyle},
  title=#2,
  coltitle=white,
  colbacktitle=black,
  fonttitle=\bfseries,
  attach boxed title to top left={xshift=2mm,yshift=-2mm},
  boxed title style={sharp corners},
  #1
}

\usepackage{amssymb}
\usepackage{amsmath}
\usepackage{lineno}
\usepackage{tikz}
\usetikzlibrary{shapes.geometric, arrows.meta, positioning, 
                 fit, backgrounds, calc, decorations.pathreplacing, backgrounds}
\usepackage[normalem]{ulem}
\usepackage[T1]{fontenc}
\usepackage{listingsutf8}
\usepackage{booktabs}
\usepackage[table]{xcolor}

\definecolor{darkblue}{rgb}{0, 0, 0.5}
\hypersetup{colorlinks=true, citecolor=darkblue, linkcolor=darkblue, urlcolor=darkblue}

\title{MedAction: Towards Active Multi-turn Clinical Diagnostic LLMs}

\usepackage[bottom]{footmisc}

\newcommand{\eqcontrib}[0]{\textsuperscript{*}}
\newcommand{\cosenior}[0]{\textsuperscript{\dag}}
\newcommand{\authorsep}[0]{\ \ }
\author{
\vspace{-1em}\\
\textbf{Hsin-Ling Hsu\textsuperscript{1}} \authorsep
\textbf{Zizheng Wang\textsuperscript{2}\eqcontrib} \authorsep
\textbf{Donghua Zhang\textsuperscript{3}\eqcontrib} \authorsep
\textbf{Nai-Chia Chen\textsuperscript{1}} \authorsep
\textbf{Jerry Wang\textsuperscript{1}}
\\
\textbf{Jun-En Ding\textsuperscript{4}} \authorsep
\textbf{Chia-Hsuan Hsu\textsuperscript{5}} \authorsep
\textbf{Guoan Wang\textsuperscript{4}} \authorsep
\textbf{Feng Liu\textsuperscript{6}} \authorsep
\textbf{Fang-Ming Hung\textsuperscript{7}}
\\
\textbf{Chenwei Wu\textsuperscript{3}\cosenior} \authorsep
\textbf{Liyue Shen\textsuperscript{3}\cosenior}
\\[6pt]
\textsuperscript{1}National Chengchi University \quad
\textsuperscript{2}Georgetown University \quad
\textsuperscript{3}University of Michigan \\
\textsuperscript{4}Stevens Institute of Technology \quad
\textsuperscript{5}National Taiwan University of Science and \\Technology \quad
\textsuperscript{6}Rutgers University \quad
\textsuperscript{7}Far Eastern Memorial Hospital
}

\begin{document}

\ifcolmsubmission
\linenumbers
\fi

\maketitle

\renewcommand{\thefootnote}{\fnsymbol{footnote}}
\footnotetext[1]{These authors contributed equally.}
\footnotetext[2]{Co-senior authorship. Contact: \href{mailto:chenweiwu99@gmail.com}{\texttt{chenweiwu99@gmail.com}}}

\begin{abstract}
% The abstract paragraph should be indented 1/2~inch (3~picas) on both left and
% right-hand margins. Use 10~point type, with a vertical spacing of 11~points.
% The word \textit{Abstract} must be centered and in point size 12. Two
% line spaces precede the abstract. The abstract must be limited to one
% paragraph.
Most existing LLM diagnoses are evaluated on static, single-turn settings where complete patient information is provided upfront, an oversimplification of real clinical practice. We study active diagnosis: the real-life clinical process of starting from initial observation, ordering tests, interpreting results, and updating a differential diagnosis across multiple turns. Through systematic analysis, we identify three recurring failure modes in current LLMs --- ungrounded test ordering, unreliable diagnostic update, and degraded multi-turn coherence --- that together reveal a core deficit: existing medical training data teaches models to reason from complete information but not to act under evolving, partial evidence. To address this gap, we introduce MedAction, a tree-structured distillation pipeline that synthesizes diverse and high-quality multi-turn diagnostic trajectories via LLM--environment interaction. We propose two knowledge-graph-grounded metrics to filter trajectory quality: Disease Trajectory Consistency (DTC), which tracks whether the model's hypothesis converges toward the correct diagnosis, and Reasoning–Action Consistency (RAC), which verifies that belief updates are driven by gathered evidence. Using this pipeline, we construct MedAction-32K, a dataset of 32,681 trajectories from 2,896 PMC cases. Fine-tuning an 8B model on MedAction-32K achieves state-of-the-art performance among open-source models on both MedR-Bench and our curated MedAction-300-Hard benchmark, pushing the edge for open-source medical LLMs.\renewcommand{\thefootnote}{\arabic{footnote}}\footnotemark
\end{abstract}
\renewcommand{\thefootnote}{\arabic{footnote}}
\footnotetext{Code and data available at \url{https://github.com/JustinHsu1019/MedAction}}

\section{Introduction}

Large language models (LLMs) have shown promising performance on medical 
question answering (QA) benchmarks, suggesting great potential 
as clinical decision-support tools for diagnosis~\citep{singhal2023large,jin2021disease}. 
However, most existing 
benchmarks formulate the medical diagnosis as a static, one-shot QA task: 
the model receives a complete and fully assembled patient record including laboratory results, imaging, and biomarkers, implicitly assuming such data are already available, and then is asked to predict the diagnosis via a well-posed question prompt. 
This setup is highly simplified and unrealistic and does not 
reflect the complexities of real-world clinical diagnosis.

Instead, as shown in Figure~\ref{fig:overview}, medical diagnosis is an active, iterative 
process. 
In practice, clinicians 
% never receive a complete picture upfront. Instead, they 
always begin with incomplete information,
then iteratively recommend new tests 
to resolve uncertainty, 
and progressively narrow the differential diagnosis based on new test results accordingly, 
until making a final definitive diagnosis with sufficient confidence. 
We call this new formulation \textit{active diagnosis 
with iterative test ordering}. 
To build LLMs that can be genuinely useful in clinical settings, 
we must enable their capability for active diagnosis via evaluation in such multi-turn, information-gathering regime.

\begin{figure}
    \centering
    \includegraphics[width=1\linewidth]{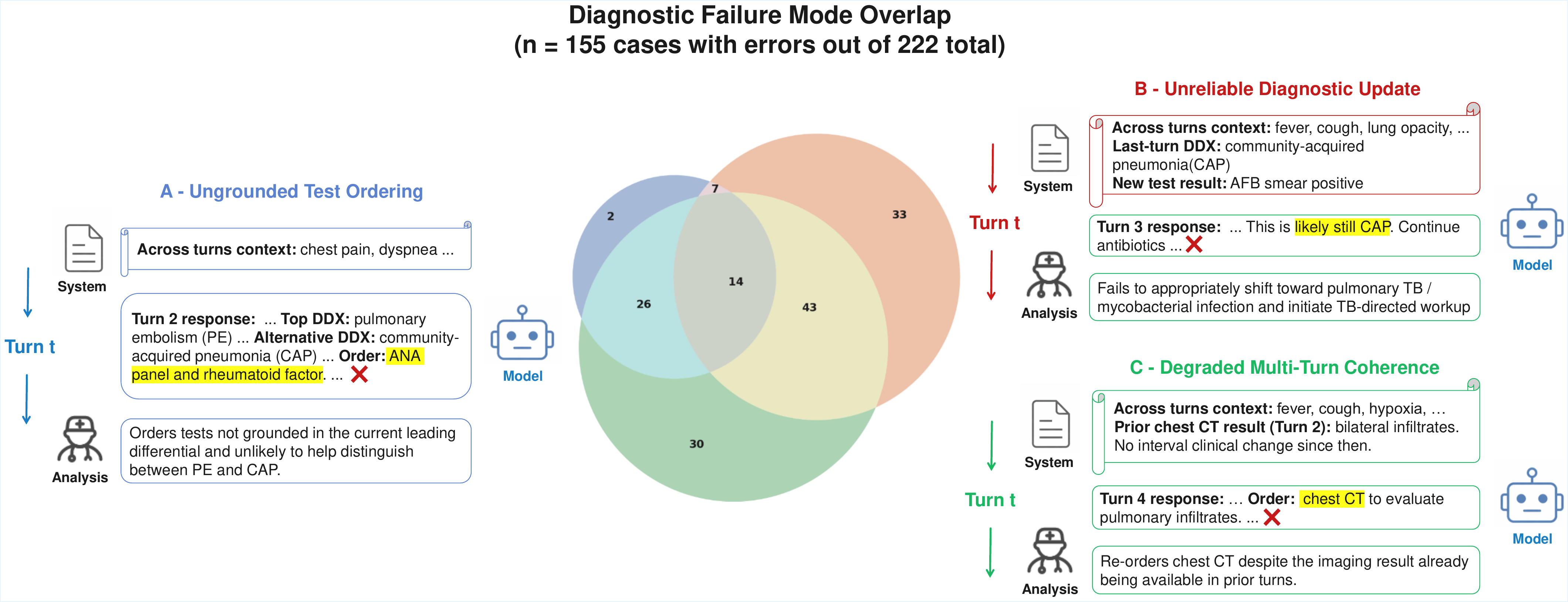}
    \caption{\textbf{Three representative failure modes (MedGemma) in active diagnosis.} Each panel illustrates one failure mode with a concrete example: (A)~ungrounded test ordering, (B)~unreliable diagnostic update, and (C)~degraded multi-turn coherence. The Venn diagram shows failure mode co-occurrence across 155 erroneous cases out of 222 examined cases in our training set.}
    \label{fig:failure_mode}
    \vspace{-10pt}
\end{figure}

While state-of-the-art LLMs perform well on static medical QA~\citep{jin2021disease,pal2022medmcqa,jin2019pubmedqa} with growing ability in multi-turn tasks in general domains~\citep{wu2025collabllm,liu2024agentbench,yao2025taubench}, a recent study~\citep{qiu2025quantifying} reveals a striking performance degradation when the same models are evaluated on active medical diagnosis task. 
This gap persists even for powerful agentic models such as GPT-o3~\citep{openai2025o3mini} and medical specialist models such as MedGemma~\citep{sellergren2025medgemma}. To understand what makes this setting uniquely challenging, we investigate LLM behavior in active multi-turn diagnosis and identify three representative failure modes (Figure~\ref{fig:failure_mode}).

\noindent\textbf{Ungrounded test ordering.} 
The clinical action space spans thousands of laboratory tests, imaging studies, and procedures, resulting in a highly complex decision landscape.
% far larger and less structured than the tool menus in general-domain agentic settings. 
We find that LLMs frequently order tests without a clear connection to their stated hypotheses, lacking case-specific clinical reasoning and context-grounded actions.
% defaulting to common symptom--test associations rather than grounding actions in individual case context.

\noindent\textbf{Unreliable diagnostic update.} 
Interpretation of clinical test results is non-trivial, requiring consideration of competing hypotheses, prior evidence, and patient context. 
Due to imprecise understanding of newly acquired test results, LLMs often adhere rigidly to an initial diagnosis despite contradictory evidence, or shift erratically between unrelated diagnoses.

\noindent\textbf{Degraded multi-turn coherence.} 
As evidence accumulates across multi-turn interactions, LLMs often lose track of which tests have been ordered, what results were obtained, and what uncertainty remains unresolved, leading to inconsistencies in longer context.

These failures reveal critical limitations of existing medical LLMs, which are usually finetuned with medical training data such as chain-of-thought (CoT) reasoning trajectories~\citep{wu2025medreason,chen-etal-2025-towards-medical}: 
models are trained to \emph{think} --- perform reasoning toward a diagnosis assuming complete information --- but not to \emph{act} --- sequentially decide which new information to gather (``action'') and update the current hypothesis (``status'') in response to newly collected information. 
Addressing these challenges in active diagnosis requires the development of multi-turn interactive medical training data that can equip LLMs with such abilities to both reason and act correctly.
% Capturing both is essential for active diagnosis, yet such high-quality data remains scarce.

% \begin{figure*}[t]
%   \centering
%   \input{fig_overview.tex}
%   \caption{\textbf{Static vs.\ Active Diagnosis.} \textit{Left:} Existing benchmarks provide a complete patient record and ask for a one-shot diagnosis. \textit{Right:} In active diagnosis with iterative test ordering, the model begins with only the chief complaint (CC) and patient history, orders tests to resolve uncertainty, and progressively narrows the differential diagnosis across multiple turns until a definitive diagnosis is reached.}
%   \label{fig:overview}
% \end{figure*}

\begin{figure*}[t]
  \centering
  \begin{tikzpicture}[
    scale=0.88, transform shape,
    >=Stealth,
    arrowmain/.style={->, very thick, color=black!50},
]

% =============================================
% LEFT: Static Diagnosis
% =============================================
\node[font=\normalsize\bfseries, color=red!60!black] at (2.6, 3.0) 
    {Static Diagnosis};

\node[draw, very thick, rounded corners=4pt, fill=gray!5, draw=gray!40,
      minimum width=4.2cm, minimum height=2.4cm] (record) at (2.6, 1.0) {};
\node[font=\scriptsize\bfseries, color=black!50] at (2.6, 1.95) 
    {Complete Patient Record};

\foreach \x/\y/\c/\t in {
    1.4/1.35/blue!12/{CC},
    3.8/1.35/green!12/{Labs},
    1.4/0.75/blue!12/{Exam},
    3.8/0.75/orange!12/{Imaging},
    1.4/0.15/blue!12/{Vitals},
    3.8/0.15/purple!12/{Pathology}} {
    \node[draw, rounded corners=2pt, thick, fill=\c, 
          minimum width=1.5cm, minimum height=0.38cm, font=\tiny, inner sep=2pt] 
        at (\x, \y) {\t};
}

\draw[arrowmain, line width=2pt] (2.6, -0.35) -- (2.6, -1.1);
\node[draw, very thick, rounded corners=3pt, fill=red!6, draw=red!45,
      font=\small\bfseries, text=red!60!black, inner sep=5pt] at (2.6, -1.55) {Diagnosis};

% =============================================
% DIVIDER
% =============================================
\draw[dashed, thick, color=gray!30] (5.8, 3.2) -- (5.8, -2.0);
\node[font=\scriptsize\bfseries, color=gray!45] at (5.8, 3.45) {vs.};

% =============================================
% RIGHT: Active Diagnosis
% =============================================
\node[font=\normalsize\bfseries, color=green!50!black] at (10.2, 3.0) 
    {Active Diagnosis};

\begin{scope}[xshift=6.6cm]

    \def\xL{0.5}
    \def\xTip{6.8}
    \def\hL{1.30}
    \def\fy{-0.1}
    \def\xBl{2.1}   % Turn 1 right edge
    \def\gap{0.3}    % gap between segments
    \def\xBr{2.4}   % Turn 2 left edge = xBl + gap
    \def\xCl{3.9}   % Turn 2 right edge
    \def\xCr{4.2}   % Turn 3 left edge = xCl + gap
    \def\xB{2.3}
    \def\xC{4.1}
    \def\xD{5.8}    % Turn 3 right edge
    \def\hB{0.951}
    \def\hC{0.601}
    \def\hD{0.272}

    % --- Fill segments (with gaps) ---
    % Segment 1
    \fill[blue!10]
        (\xL, \hL+\fy) -- (\xBl, 0.97+\fy) -- (\xBl, -0.97+\fy) -- (\xL, -\hL+\fy) -- cycle;
    % Segment 2
    \fill[green!8]
        (\xBr, 0.93+\fy) -- (\xCl, 0.63+\fy) -- (\xCl, -0.63+\fy) -- (\xBr, -0.93+\fy) -- cycle;
    % Segment 3
    \fill[orange!8]
        (\xCr, 0.59+\fy) -- (\xD, 0.272+\fy) -- (\xD, -0.272+\fy) -- (\xCr, -0.59+\fy) -- cycle;

    % --- Outline (closed trapezoids with gaps) ---
    % Segment 1
    \draw[color=gray!40, thick]
        (\xL, \hL+\fy) -- (\xBl, 0.97+\fy) -- (\xBl, -0.97+\fy) -- (\xL, -\hL+\fy) -- cycle;
    % Segment 2
    \draw[color=gray!40, thick]
        (\xBr, 0.93+\fy) -- (\xCl, 0.63+\fy) -- (\xCl, -0.63+\fy) -- (\xBr, -0.93+\fy) -- cycle;
    % Segment 3 (no tip triangle)
    \draw[color=gray!40, thick]
        (\xCr, 0.59+\fy) -- (\xD, 0.272+\fy) -- (\xD, -0.272+\fy) -- (\xCr, -0.59+\fy) -- cycle;
    
    % % --- Labels ---
    % \node[font=\scriptsize\bfseries, color=blue!60!black] 
    %     at ({(\xL+\xBl)/2}, \fy+0.12) {DDx};
    % \node[font=\tiny, color=blue!40] 
    %     at ({(\xL+\xBl)/2}, \fy-0.25) {6 Hs};
    % \node[font=\scriptsize\bfseries, color=green!50!black] 
    %     at ({(\xBr+\xCl)/2}, \fy+0.12) {DDx};
    % \node[font=\tiny, color=green!35] 
    %     at ({(\xBr+\xCl)/2}, \fy-0.2) {4 Hs};
    % \node[font=\scriptsize\bfseries, color=orange!60!black] 
    %     at ({(\xCr+\xD)/2}, \fy+0.12) {DDx};
    % \node[font=\tiny, color=orange!40] 
    %     at ({(\xCr+\xD)/2}, \fy-0.1) {2 Hs};

    % --- Labels: merged DDx + count into single label ---
    \node[font=\scriptsize\bfseries, color=blue!70!black] 
        at ({(\xL+\xBl)/2}, \fy) {6 DDx};
    \node[font=\scriptsize\bfseries, color=green!60!black] 
        at ({(\xBr+\xCl)/2}, \fy) {4 DDx};
    \node[font=\scriptsize\bfseries, color=orange!70!black] 
        at ({(\xCr+\xD)/2}, \fy) {2 DDx};
        
    % --- Turn labels ---
    \node[font=\scriptsize\bfseries, color=black!55] 
        at ({(\xL+\xB)/2}, 2.45) {Turn 1};
    \node[font=\scriptsize\bfseries, color=black!55] 
        at ({(\xB+\xC)/2}, 2.45) {Turn 2};
    \node[font=\scriptsize\bfseries, color=black!55] 
        at ({(\xC+\xD)/2}, 2.45) {Turn 3};
    \node[font=\scriptsize\bfseries, color=black!55] 
        at (\xTip, 2.45) {Final};

    % --- Evidence tags (cumulative stacking, growing downward) ---
    % Turn 1: CC + Exam
    \node[draw, rounded corners=2pt, thick, fill=blue!10, 
          font=\tiny, inner sep=2pt, minimum width=1.2cm] (ev1a)
        at ({(\xL+\xB)/2}, 1.95) {CC + Exam};
    \draw[->, thick, color=gray!45] (ev1a.south) -- ({(\xL+\xB)/2}, \hL+\fy-0.06);

    % Turn 2: CC + Exam (top, same position) + Labs (below, new)
    \node[draw, rounded corners=2pt, thick, fill=blue!10, 
          font=\tiny, inner sep=2pt, minimum width=1.0cm] (ev2a)
        at ({(\xB+\xC)/2}, 1.95) {CC + Exam};
    \node[draw, rounded corners=2pt, thick, fill=green!10, 
          font=\tiny, inner sep=2pt, minimum width=1.0cm] (ev2b)
        at ({(\xB+\xC)/2}, 1.5) {Labs};
    \draw[->, thick, color=gray!45] (ev2b.south) -- ({(\xB+\xC)/2}, \hB+\fy-0.06);

    % Turn 3: CC + Exam (top) + Labs (middle) + Imaging (bottom, new)
    \node[draw, rounded corners=2pt, thick, fill=blue!10, 
          font=\tiny, inner sep=2pt, minimum width=1.0cm] (ev3a)
        at ({(\xC+\xD)/2}, 1.95) {CC + Exam};
    \node[draw, rounded corners=2pt, thick, fill=green!10, 
          font=\tiny, inner sep=2pt, minimum width=1.0cm] (ev3b)
        at ({(\xC+\xD)/2}, 1.5) {Labs};
    \node[draw, rounded corners=2pt, thick, fill=orange!10, 
          font=\tiny, inner sep=2pt, minimum width=1.0cm] (ev3c)
        at ({(\xC+\xD)/2}, 1.05) {Imaging};
    \draw[->, thick, color=gray!45] (ev3c.south) -- ({(\xC+\xD)/2}, \hC+\fy-0.06);

    % --- Curved arrows below funnel: "Order next test" ---
    \draw[->, thick, color=orange!70!black, bend right=35] 
        ({(\xL+\xB)/2}, -\hL+\fy-0.1) to ({(\xB+\xC)/2+0.1}, -\hB+\fy-0.1);
    \draw[->, thick, color=orange!70!black, bend right=35] 
        ({(\xB+\xC)/2+0.3}, -\hB+\fy-0.1) to ({(\xC+\xD)/2}, -\hC+\fy-0.1);
    \draw[->, thick, color=orange!70!black, bend right=35] 
        ({(\xC+\xD)/2+0.3}, -\hC+\fy-0.1) to (6.6, -0.35);
    \node[font=\tiny\bfseries, color=orange!70!black] 
        at ({(\xB+\xD)/2}, -1.8) {Order next test};
\end{scope}

\node[draw, thick, rounded corners=2pt, 
      fill=red!6, draw=red!45,
      font=\scriptsize\bfseries, text=red!60!black, inner sep=4pt,
      minimum width=0.8cm, minimum height=0.5cm] 
    at (13.4, -0.1) {Diagnosis};

\end{tikzpicture}
  \caption{\textbf{Static vs.\ Active Diagnosis.} \textit{Left:} Existing benchmarks provide a complete patient record and ask for a one-shot diagnosis. \textit{Right:} In active diagnosis, the model begins with only the chief complaint (CC) and physical examination, then iteratively orders new tests to resolve uncertainty and progressively narrows the \textit{differential diagnosis (DDx)} from six candidate conditions down to a definitive diagnosis through multiple turns.}
  \label{fig:overview}
\end{figure*}
In this work, we introduce \textbf{MedAction}, a framework for automatically constructing high-quality multi-turn active diagnosis training data from publicly available medical literature. 
Unlike existing medical CoT data that teaches models to \textit{think} from complete information, MedAction trajectories capture both \textit{thinking} and \textit{acting} under partial, evolving evidence. Starting from raw case reports, we first extract 2896 interactive clinical environments that simulate the partial-information setting of active diagnosis: the model observes only the initial patient information and needs to take proactive actions to request new test results turn by turn. We then construct structured multi-turn trajectories by having state-of-the-art LLMs interact with these environments, producing chains of reasoning-action data that mirror real clinicians' diagnostic procedure: maintaining a differential, identifying diagnostic uncertainty, and selecting tests to resolve it. 

To distill high-quality training data and maximize data utilization, we propose two knowledge-graph-grounded metrics to enable clinically grounded rejection sampling. \textbf{Disease Trajectory Consistency (DTC)} measures the knowledge-graph distance between the model's current top hypothesis and the ground-truth diagnosis on PrimeKG~\citep{chandak2023building}, allowing us to retain complete trajectories that reach the correct answer as well as partially correct trajectories where the hypothesis was still converging. \textbf{Reasoning--Action Consistency (RAC)} then measures whether ordered tests drove meaningful changes in the differential on ClinGraph~\citep{johnson2024unified}, filtering individual turns where hypothesis shifts were unrelated to gathered evidence. We further investigate key design choices in trajectory construction, including teacher model selection, tree sampling strategy, and filtering thresholds, providing actionable guidance for building multi-turn medical reasoning-action data. Training on the resulting \textbf{MedAction-32K dataset}, our 8B-parameter model achieves state-of-the-art performance among open-source models under multi-turn active diagnosis settings, demonstrating that high-quality trajectory data, even at 
modest scale, drives larger gains than architecture scaling alone. In addition to the training set, we curate \textbf{MedAction-300 Hard}, a test set of 300 rare-disease cases that demand longer diagnostic trajectories and more extensive test ordering than existing benchmarks.

Our contributions are threefold:
\begin{enumerate}
\item \textbf{New formulation and failure analysis.} 
We reformulate the clinical diagnosis as multi-turn active diagnostic setting with iterative test ordering, and identify three representative failure modes that explain why current LLMs struggle in this setting.

% \item \textbf{New dataset construction and new metrics.} We introduce a tree-structured data distillation pipeline that automatically synthesizes high-quality multi-turn medical diagnostic trajectories from public case reports, investigate key design choices in trajectory construction, and release curated training and testing datasets to the community. 
% We also propose two new metrics for training-data distillation and rigorous model evaluation.

\item \textbf{New dataset construction and new metrics.} We introduce a tree-structured data distillation pipeline that automatically synthesizes high-quality multi-turn medical diagnostic trajectories from public case reports, investigate key design choices in trajectory construction, and release the MedAction-32K training set and the MedAction-300 Hard test set---a challenging benchmark of rare-disease cases---to the community. We also propose two new metrics for training-data distillation and rigorous model evaluation.

\item \textbf{Efficient and effective model.} Our dataset trains an 8B-parameter model that achieves state-of-the-art results amongst open-source models on active diagnosis and test recommendation. 
\end{enumerate}

\section{Related Work}

\begin{table}[t]
\centering
\resizebox{\textwidth}{!}{%
\begin{tabular}{l|ccccc|c}
\toprule
\textbf{Feature} 
& \makecell{\textbf{Static} \\ \textbf{Benchmarks} \\ \small \citep{jin2021disease} \\ \small \citep{jin2019pubmedqa}}
& \makecell{\textbf{Interactive} \\ \textbf{Inquiry} \\ \small \citep{li2024mediq} \\ \small \citep{lai2026doctorr}}
& \makecell{\textbf{Multi-Agent} \\ \textbf{Simulation} \\ \small \citep{schmidgall2024agentclinic}} 
& \makecell{\textbf{Multi-turn} \\ \textbf{Eval} \\ \small \citep{qiu2025quantifying}} 
& \makecell{\textbf{Training Data} \\ \textbf{Construction} \\ \small \citep{wu2025medreason} \\ \small \citep{ossowski2025octomed}}
& \makecell{\textbf{MedAction} \\ \textbf{(Ours)}} \\
\midrule
Multi-turn diagnosis & \ding{55} & \ding{51} & \ding{51} & \ding{51} & \ding{55} & \ding{51} \\
Iterative test ordering & \ding{55} & \ding{55} & \ding{51} & \ding{51} & \ding{55} & \ding{51} \\
Interactive data generation & \ding{55} & \ding{55} & \ding{55} & \ding{55} & \ding{55} & \ding{51} \\
Turn-level quality filtering & \ding{55} & \ding{55} & \ding{55} & \ding{55} & \ding{55} & \ding{51} \\
Data recipe & \ding{55} & \ding{51} & \ding{55} & \ding{55} & \ding{51} & \ding{51} \\
Evaluation benchmark & \ding{51} & \ding{51} & \ding{51} & \ding{51} & \ding{55} & \ding{51} \\
\bottomrule
\end{tabular}%
}
\caption{Comparison with existing medical benchmarks and training frameworks.}
\label{tab:comparison}
\vspace{2pt}
\end{table}

We summarize the comparison with prior work in Table~\ref{tab:comparison}.

\textbf{Beyond Static Diagnosis LLM Benchmarks}
Standard medical benchmarks such as MedQA~\citep{jin2021disease}, MedMCQA~\citep{pal2022medmcqa}, 
and PubMedQA~\citep{jin2019pubmedqa} evaluate LLMs in a static, single-turn format with complete records. Recent work has begun to study more realistic settings: MediQ~\citep{li2024mediq} converts static MedQA cases into partial-information 
scenarios requiring models to ask follow-up questions; MedR-Bench~\citep{qiu2025quantifying} evaluates LLMs on test recommendation and diagnosis over multiple turns, finding that accuracy drops sharply on examination recommendation even for models performing well on static tasks. Other efforts include MedJourney~\citep{wu2024medjourney}, which decomposes the clinical process into sequential stages; AMIE~\citep{tu2025towards}, which studies conversational diagnostic AI; and Liu et al.~\citep{liu2025multiturn}, which analyze challenges in multi-turn diagnostic LLM systems. Despite this progress, the underlying causes of failure in active diagnosis, and how to construct training data for this setting, remain unexplored.

\textbf{Medical LLM Training Data Construction}
Carefully curated reasoning traces~\citep{muennighoff2025s1,guha2026openthoughts}
can significantly improve reasoning performance. Applying these ideas to medicine requires reasoning chains to be factually accurate. MedReason~\citep{wu2025medreason}
tackles this by grounding CoT generation in a medical knowledge graph; while other works distill long reasoning traces from stronger models with domain-specific verification~\citep{ossowski2025octomed,chen-etal-2025-towards-medical,huang2025m1}. UltraMedical~\citep{zhang2024ultramedical} provides a large-scale biomedical training resource. However, all existing medical data construction work targets
single-turn reasoning from complete information, neglecting the real-life active diagnosis setup. MedAction directly addresses this gap by
constructing trajectories that capture both the reasoning and actions at each diagnostic step.

\section{Methodology}
\label{sec:methodology}

\subsection{Problem Formulation}

We define \textbf{active diagnosis with iterative test ordering} as a 
sequential decision-making problem under partial observability. Let $s^*$ denote the ground-truth diagnosis. A diagnostic trajectory $\tau = (o_1, a_1, r_1, o_2, a_2, r_2, \dots, o_T)$ consists of: \textbf{observations} $o_t$, the accumulated clinical evidence at turn $t$ (initial presentation plus all prior test results), where partial observability arises because only results of ordered tests are revealed; \textbf{actions} $a_t \subseteq \mathcal{A}$, clinical tests selected from a large open-vocabulary action space; \textbf{results} $r_t$, the findings returned for $a_t$ (lab values, imaging reports, or clinical observations); and \textbf{belief state} $b_t$, the model's ranked differential diagnosis approximating the posterior over $s^*$. The process terminates when the model commits to a definitive diagnosis or the turn limit $T_{\max}$ is reached.

\begin{figure*}[t]
  \centering
  \begin{tikzpicture}[
    scale=0.88, transform shape,
    >=Stealth,
    node distance=0.4cm,
    % --- color definitions ---
    casecolor/.style={fill=orange!8, draw=orange!60, thick, rounded corners=3pt},
    envcolor/.style={fill=blue!6, draw=blue!50, thick, rounded corners=3pt},
    treecolor/.style={fill=green!6, draw=green!50!black, thick, rounded corners=3pt},
    % --- node styles ---
    component/.style={draw, thick, rounded corners=2pt, minimum height=0.7cm, 
                      font=\small, inner sep=4pt},
    smallbox/.style={draw, rounded corners=1pt, font=\scriptsize, inner sep=3pt,
                     minimum height=0.5cm, minimum width=1.2cm},
    % --- tree node styles ---
    tnode/.style={circle, draw, thick, minimum size=0.35cm, inner sep=0pt},
    tnode_good/.style={tnode, fill=green!25, draw=green!60!black},
    tnode_bad/.style={tnode, fill=red!15, draw=red!50},
    tnode_trunc/.style={tnode, fill=yellow!25, draw=yellow!70!black},
    tnode_neutral/.style={tnode, fill=gray!15, draw=gray!60},
    % --- arrow styles ---
    stage_arrow/.style={->, line width=1.8pt, color=black!70},
    interact_arrow/.style={->, thick, color=blue!60, dashed},
    tree_edge/.style={-, thick, color=gray!70},
    tree_edge_pruned/.style={-, thick, color=red!30, dashed},
]

% =============================================
% PANEL 1: Case Report (left)
% =============================================
\begin{scope}[local bounding box=case_panel]
    % Document shape
    \node[draw, thick, minimum width=2.2cm, minimum height=3.2cm, 
          fill=orange!5, rounded corners=2pt] (doc) at (0, 0) {};
    % Mixed content lines (representing retrospective, interleaved info)
    \foreach \y/\w/\c in {
        0.9/1.6/blue!40, 0.6/1.4/blue!40, 0.3/1.7/blue!40,
        0.0/1.0/red!40, -0.2/1.5/red!40,
        -0.5/1.3/blue!40, -0.7/0.9/red!40, -0.9/1.6/blue!40} {
        \fill[\c, rounded corners=0.5pt] ([xshift=-0.8cm, yshift=\y cm]doc.center) 
            rectangle +(\w cm, 0.12cm);
    }
    % Label
    \node[below=0.3cm of doc, font=\small\bfseries] (case_label) {Case Report};
    \node[below=0.0cm of case_label, font=\scriptsize, color=gray!70] {(PMC)};
\end{scope}

% =============================================
% STAGE 1 ARROW
% =============================================
\draw[stage_arrow] ([xshift=0.3cm]doc.east) -- ++(1.4, 0)
    node[midway, above, font=\scriptsize\bfseries, color=black!70] {Stage 1};

% =============================================
% PANEL 2: Clinical Environment (middle)
% =============================================
\begin{scope}[local bounding box=env_panel, xshift=5.0cm]
    % Background box
    \node[envcolor, minimum width=3.8cm, minimum height=4.0cm] (env_bg) at (0, 0) {};
    \node[above=0.05cm of env_bg.north, font=\small\bfseries, color=blue!70!black] 
        {};
    
    % o1: Initial Observation
    \node[smallbox, fill=blue!12, minimum width=3.2cm] (obs) at (0, 1.2) 
        {$o_1$: Initial Observation};
    
    % M: Test Menu
    \node[smallbox, fill=blue!12, minimum width=3.2cm] (menu) at (0, 0.3) 
        {$\mathcal{M}$: Test Menu};
    % Small test items
    \node[draw, rounded corners=1pt, font=\tiny, fill=white, inner sep=2pt] 
        (t1) at (-0.9, -0.25) {CBC};
    \node[draw, rounded corners=1pt, font=\tiny, fill=white, inner sep=2pt] 
        (t2) at (-0.15, -0.25) {CT};
    \node[draw, rounded corners=1pt, font=\tiny, fill=white, inner sep=2pt] 
        (t3) at (0.5, -0.25) {MRI};
    \node[draw, rounded corners=1pt, font=\tiny, fill=white, inner sep=2pt] 
        (t4) at (1.1, -0.25) {$\cdots$};
    
    % f: Oracle Function
    \node[smallbox, fill=blue!12, minimum width=3.2cm] (oracle) at (0, -1.0) 
        {$f$: Oracle Function};
    \node[font=\tiny, color=gray!70, text width=3cm, align=center] at (0, -1.55) 
        {$a \in \mathcal{M} \to$ result \quad $a \notin \mathcal{M} \to$ \texttt{UNAVAIL.}};
\end{scope}

% =============================================
% INTERACTION LOOP (between env and tree)
% =============================================
% Label above
\node[font=\tiny, text width=1.2cm, align=center, color=blue!60] 
    at (8.0, 1.6) {Multi-turn\\interaction};
% a_t arrow: agent -> env (above Stage 2 arrow, going left)
\draw[interact_arrow, bend right=25] 
    (8.5, 0.8) to 
    node[above, font=\tiny, color=blue!50] {$a_t$}
    (7.5, 0.8);
% r_t arrow: env -> agent (below Stage 2 arrow, going right)
\draw[interact_arrow, bend right=25] 
    (7.5, -0.4) to 
    node[below, font=\tiny, color=blue!50] {$r_t$}
    (8.5, -0.4);
% =============================================
% STAGE 2 ARROW
% =============================================
\draw[stage_arrow] ([xshift=0.3cm]env_bg.east) -- ++(1.6, 0)
    node[midway, above, font=\scriptsize\bfseries, color=black!70] {Stage 2};

\begin{scope}[local bounding box=tree_panel, xshift=9.5cm]
    
    % Turn labels on the right
    \node[font=\tiny, color=gray!50] at (4.2, 1.6) {turn 1};
    \node[font=\tiny, color=gray!50] at (4.2, 0.5) {turn 2};
    \node[font=\tiny, color=gray!50] at (4.2, -0.6) {turn 3};
    \node[font=\tiny, color=gray!50] at (4.2, -1.7) {turn 4};
    \node[font=\tiny, color=gray!50] at (4.2, -2.8) {turn 5};
    
    % --- ROOT (turn 1) ---
    \node[tnode_neutral, minimum size=0.45cm, font=\tiny] (root) at (0, 1.6) {$o_1$};
    
    % --- TURN 2: 3 branches from root ---
    \node[tnode_good] (t2a) at (0, 0.5) {};
    \node[tnode_good] (t2b) at (0.8, 0.5) {};
    \node[tnode_bad]  (t2c) at (3.2, 0.5) {};
    
    \draw[tree_edge] (root) -- (t2a);
    \draw[tree_edge] (root) -- (t2b);
    \draw[tree_edge] (root) -- (t2c);
    
    % --- TURN 3 ---
    % Left branch: straight down
    \node[tnode_good] (t3a) at (0, -0.6) {};
    \draw[tree_edge] (t2a) -- (t3a);
    
    % Middle branch: fans out, first child aligns vertically with t2b
    \node[tnode_good] (t3b1) at (0.8, -0.6) {};
    \node[tnode_good] (t3b2) at (1.6, -1.7) {};
    \node[tnode_good] (t3b3) at (2.4, -0.6) {};
    \draw[tree_edge] (t2b) -- (t3b1);
    \draw[tree_edge] (t3b1) -- (t3b2);
    \draw[tree_edge] (t2b) -- (t3b3);
    % Scissors: truncation starts here
    % \draw[yellow!70!black, thick, dashed] 
    %     ([xshift=-0.3cm, yshift=0.25cm]t3b3.north west) -- 
    %     ([xshift=0.3cm, yshift=-0.25cm]t3b3.south east);
    
    % Right branch: rejected
    \node[tnode_bad] (t3c) at (3.2, -0.6) {};
    \draw[tree_edge_pruned] (t2c) -- (t3c);
    % Scissors: cut starts here
    \draw[red!50, thick, dashed] 
        ([xshift=-0.3cm, yshift=0.25cm]t2c.north west) -- 
        ([xshift=0.3cm, yshift=-0.25cm]t2c.south east);
    
    % --- TURN 4 ---
    % Left: straight down
    \node[tnode_good] (t4a) at (0, -1.7) {};
    \draw[tree_edge] (t3a) -- (t4a);
    
    % Middle-left: t3b1 -> one child, aligns vertically
    \node[tnode_good] (t4b1) at (0.8, -1.7) {};
    \draw[tree_edge] (t3b1) -- (t4b1);
    
    % Convergent prefix: continues down vertically
    \node[tnode_trunc] (t4b4) at (2.4, -1.7) {};
    \draw[tree_edge, color=yellow!70!black, densely dotted] (t3b3) -- (t4b4);
    \draw[yellow!70!black, thick, dashed] 
        ([xshift=-0.3cm, yshift=0.25cm]t4b4.north west) -- 
        ([xshift=0.3cm, yshift=-0.25cm]t4b4.south east);
    
    % Right: rejected continues vertically
    \node[tnode_bad] (t4c) at (3.2, -1.7) {};
    \draw[tree_edge_pruned] (t3c) -- (t4c);
    
    % --- TURN 5 (leaf nodes) ---
    % Left: straight green
    \node[tnode_good, label={[font=\tiny, color=green!50!black]below:$\checkmark$}] 
        (t5a) at (0, -2.8) {};
    \draw[tree_edge] (t4a) -- (t5a);
    
    % Middle leaves
    \node[tnode_good, label={[font=\tiny, color=green!50!black]below:$\checkmark$}] 
        (t5b1) at (0.8, -2.8) {};
    \node[tnode_good, label={[font=\tiny, color=green!50!black]below:$\checkmark$}] 
        (t5b3) at (1.6, -2.8) {};
    \draw[tree_edge] (t4b1) -- (t5b1);
    \draw[tree_edge] (t3b2) -- (t5b3);
    
    % Convergent prefix leaf
    \node[tnode_trunc, label={[font=\tiny, color=yellow!70!black]below:{\tiny prune}}] 
        (t5b4) at (2.4, -2.8) {};
    \draw[tree_edge, color=yellow!70!black, densely dotted] (t4b4) -- (t5b4);
    
    % Right: rejected leaf
    \node[tnode_bad, label={[font=\tiny, color=red!50]below:$\times$}] 
        (t5c) at (3.2, -2.8) {};
    \draw[tree_edge_pruned] (t4c) -- (t5c);
\end{scope}

% =============================================
% STAGE 3 LABEL (below tree)
% =============================================
\node[font=\scriptsize\bfseries, color=black!70] at (10.8, -4.0) {Stage 3: Metrics \& Pruning};

% =============================================
% LEGEND (horizontal, spanning full width at bottom)
% =============================================
\begin{scope}[yshift=-4.8cm]
    % Item 1: Retained
    \node[tnode_good, minimum size=0.25cm] (leg1) at (2, 0) {};
    \node[anchor=mid west, font=\tiny] at (2.25, 0) {Retained (GT hit)};
    
    % Item 2: Truncated (partial match)
    \node[tnode_trunc, minimum size=0.25cm] (leg2) at (4.8, 0) {};
    \node[anchor=mid west, font=\tiny] at (5.05, 0) {truncated (partial)};
    
    % Item 3: Pruned (rejected)
    \node[tnode_bad, minimum size=0.25cm] (leg3) at (7.6, 0) {};
    \node[anchor=mid west, font=\tiny] at (7.85, 0) {Pruned (rejected)};
\end{scope}

% =============================================
% SECTION LABELS (top, spanning each panel)
% =============================================
\node[font=\footnotesize, color=orange!70!black] at (0, 2.8) {\textbf{Source Data}};
\node[font=\footnotesize, color=blue!70!black] at (5.0, 2.8) {\textbf{Clinical Environment $\mathcal{E}$}};
\node[font=\footnotesize, color=green!50!black] at (10.8, 2.8) {\textbf{Trajectory Tree $\mathcal{T}$}};

\end{tikzpicture}
  \caption{\textbf{Overview of the MedAction pipeline.} Stage~1 transforms 
retrospective case reports into interactive clinical environments. 
Stage~2 generates a trajectory tree through multi-turn 
LLM--environment interaction. Stage~3 applies 
knowledge-graph-grounded metrics to retain high-quality 
trajectories: complete trajectories reaching the ground-truth 
diagnosis (green, $\checkmark$), truncated prefixes from partially 
correct paths (yellow), while rejecting divergent or ungrounded 
trajectories (red, $\times$).}
  \label{fig:main_methodology}
  \vspace{-15pt}
\end{figure*}
\subsection{Overview}
\label{sec:overview}

MedAction addresses the scarcity of high-quality multi-turn diagnostic training data through a three-stage pipeline (Figure~\ref{fig:main_methodology}):
\begin{equation}
\label{eq:pipeline}
\scalebox{0.85}{$\displaystyle
    \underbrace{\text{Case Reports}}_{\text{PMC}} \;\xrightarrow{\;\text{Stage 1}\;}\; \underbrace{\text{Clinical Environment}}_{\mathcal{E}} \;\xrightarrow{\;\text{Stage 2}\;}\; \underbrace{\text{Trajectory Tree}}_{\mathcal{T}} \;\xrightarrow{\;\text{Stage 3}\;}\; \underbrace{\text{Training Data}}_{\mathcal{D}}
$}
\end{equation}

We first transform each case report into an interactive \textit{clinical environment} (Stage 1), generate tree-like diagnostic trajectories by having 
state-of-the-art LLMs interact with this environment (Stage 2), and 
filter the resulting trajectories using knowledge-graph-grounded metrics 
(Stage 3).

\subsection{Data Generation Framework}
\label{sec:data_gen}

\paragraph{Stage 1: Clinical Environment Construction}
\label{sec:env_construction}
From each PMC-OA case report~\citep{nlm2003pmcoa}, we use an LLM to extract a clinical environment $\mathcal{E} = (o_1, \mathcal{M}, f)$: the \textbf{initial observation} $o_1$ (chief complaint, history, physical exam); the \textbf{test menu} $\mathcal{M} \subseteq \mathcal{A}$ of documented tests; and an \textbf{oracle function} $f: \mathcal{A} \to \mathcal{R} \cup \{\texttt{UNAVAILABLE}\}$ returning documented results or \texttt{UNAVAILABLE}.
\paragraph{Stage 2: Multi-turn Chain-of-Reasoning-Action Trajectory Distillation}
\label{sec:distillation}

Rather than learning a black-box policy $\pi(a_t \mid o_t)$, we require 
the model to follow an explicit reason-action pattern at each turn:
\begin{equation}
\label{eq:reasoning_chain}
    o_t \;\xrightarrow{\text{analyze}}\; b_t
        \;\xrightarrow{\text{identify}}\; p_t
        \;\xrightarrow{\text{select}}\; a_t
\end{equation}
where $b_t$ is a ranked differential with confidence estimates, $p_t$ states the key remaining uncertainty distinguishing top hypotheses, and $a_t$ selects tests justified by $p_t$. The oracle returns $r_t = f(a_t)$, appended to $o_t$ for the next turn. To prevent context degradation, we maintain a cumulative record of all tests and results alongside the model's reasoning from the last two turns.

To maximize diversity, we employ tree-structured sampling: We sample $K$ independent rollouts from $o_1$ using multiple teacher LLMs, as well as branched rollouts launched from intermediate states of existing trajectories, producing diverse continuations from shared evidence states.

\paragraph{Stage 3: Metrics and Data Pruning} Not all sampled trajectories exhibit coherent diagnostic reasoning. We 
define two knowledge-graph-grounded metrics to filter them, derived from 
shortest-path distances on PrimeKG~\citep{chandak2023building} 
(disease hypothesis--disease ground truth relationships) and ClinGraph~\citep{
johnson2024unified} (test--disease relationships), with free-text 
entities mapped to graph nodes via a biomedical information retrieval model~\citep{jin2023medcpt}.

\paragraph{Disease Trajectory Consistency (DTC).}
DTC measures the distance between the model's current top hypothesis and the ground-truth diagnosis $s^*$, quantifying whether the diagnostic trajectory is converging toward or diverging from the correct answer. Using PrimeKG:
\begin{equation}
\label{eq:dtc}
    \text{DTC}_t = h_{\text{PrimeKG}}(b_t^{(1)},\; s^*)
\end{equation}
where $b_t^{(1)}$ denotes the top-ranked diagnosis in the differential diagnosis list $b_t$ at turn $t$, and $h_{\text{PrimeKG}}$ denotes the shortest-path distance on PrimeKG. A trajectory with decreasing DTC over turns indicates convergence toward the correct diagnosis; increasing DTC signals divergence. $\text{DTC}_t = 0$ means the model has identified the correct diagnosis at turn $t$. For pruning, we scan backward from the final turn $T$, searching for the latest turn $t^*$ satisfying $\text{DTC}_{t^*} \leq \text{DTC}_1$, that is, the latest turn at which the model's hypothesis is at least as close to the ground truth as its initial guess. All turns after $t^*$ are truncated. If no such turn $t^* \geq 2$ exists (i.e., every subsequent turn is strictly farther from the ground truth than the initial hypothesis), the trajectory is discarded entirely, as it demonstrates no diagnostic convergence despite multiple rounds of evidence gathering.

\paragraph{Reasoning--Action Consistency (RAC).}
RAC measures whether the tests ordered in the previous turn actually drove meaningful changes in the differential diagnosis, providing a link between actions and belief updates. Let $\Delta b_t = (b_t \setminus b_{t-1}) \cup (b_{t-1} \setminus b_t)$ denote the set of diagnoses that were added or removed between turns $t{-}1$ and $t$, identified by matching free-text diagnoses to knowledge graph node identifiers rather than surface string comparison. For each changed diagnosis $d \in \Delta b_t$, we compute its minimum shortest-path distance to any action ordered in the previous turn:
\begin{equation}
\label{eq:rac}
    \text{RAC}_t = \frac{1}{|\Delta b_t|} \sum_{d \in \Delta b_t} \min_{a' \in a_{t-1}} h_{\text{ClinGraph}}(d,\; a'), \quad t \geq 2
\end{equation}
Lower RAC indicates evidence-driven updates; higher values indicate arbitrary hypothesis shifting. For each turn $t \geq 2$ retained after DTC pruning, if $\text{RAC}_t > \tau_{\text{RAC}}$ (a maximum hop threshold), we remove the preceding turn $t{-}1$ that ordered ineffective tests. This pruning is applied in a single pass.  

\section{Experiments}

We train an II-Medical-8B~\citep{2025II-Medical-8B} model based on Qwen-3 8B~\citep{qwen3technicalreport} model using supervised fine-tuning (SFT) on MedAction-32K, which consists of 32681 samples from 2896 cases distilled using Baichuan-M3~\citep{dou2026baichuan} and GPT5.4-mini~\citep{openaigpt54mini} as our teachers. 
We evaluate on the Med-R~\citep{qiu2025quantifying} dataset and our own MedAction-300 Hard test dataset (Appendix~\ref{app:medaction300}). We compare our model against open-source general domain models including
Qwen-QwQ (32B)~\citep{qwq32b,qwen2.5}, closed-source models including
OpenAI-o3-mini~\citep{openai2025o3mini}, Gemini 2.5 Flash-Lite~\citep{comanici2025gemini},
and GPT-5.4~\citep{openai2026gpt54thinking}, and medical domain open-source models including
Baichuan-M1 (14B)~\citep{wang2025baichuan}, Baichuan-M3 (235B-8bit)~\citep{dou2026baichuan},
MedGemma (27B)~\citep{sellergren2025medgemma}, and Diagnosis-GPT (34B)~\citep{chen-etal-2025-cod}.
% We report average test recommendation precision, recall and F1, as well as multi-turn diagnosis accuracy across 3 runs. More training and inference configs are included in Appendix~\ref{app:training_config}. Prompt templates are provided in Appendix~\ref{app:prompt_template}.
We report average test recommendation precision, recall and F1, as well as multi-turn diagnosis accuracy across 3 runs. In addition, we conduct a blinded expert evaluation on a subset of MedAction-300-Hard to assess the clinical quality of diagnostic trajectories and compare expert judgments with GPT-4o evaluation. More training and inference configs are included in Appendix~\ref{app:training_config}. Prompt templates are provided in Appendix~\ref{app:prompt_template}.

\section{Results}
\begin{table*}[t]
\centering
\small
\begin{tabular}{l cccc cccc}
\toprule
& \multicolumn{4}{c}{\textbf{MedR-Bench}} & \multicolumn{4}{c}{\textbf{MedAction-300-Hard}} \\
\cmidrule(lr){2-5} \cmidrule(lr){6-9}
\textbf{Model} & Prec & Rec & F1 & Diag Acc & Prec & Rec & F1 & Diag Acc \\
\midrule
\multicolumn{9}{l}{\textit{Closed-Source Models}} \\
OpenAI-o3-mini       & \underline{0.36} & 0.42 & 0.39 & \underline{0.71} & \textbf{0.54} & 0.47 & \textbf{0.50} & \underline{0.67} \\
Gemini 2.5 Flash-Lite     & \underline{0.36} & \underline{0.44} & \underline{0.40} & 0.68 & \underline{0.47} & \underline{0.48} & 0.47 & 0.63 \\
GPT-5.4              & \textbf{0.47} & \textbf{0.56} & \textbf{0.51} & \textbf{0.81} & 0.42 & \textbf{0.60} & \underline{0.49} & \textbf{0.82} \\
\midrule
\multicolumn{9}{l}{\textit{Open-Source Models}} \\
Baichuan-M1 (14B)    & 0.42 & 0.39 & 0.40 & 0.61 & 0.47 & 0.41 & 0.44 & 0.56 \\
Baichuan-M3 (235B)   & 0.52 & \underline{0.52} & \underline{0.52} & \underline{0.69} & 0.50 & \textbf{0.54} & \textbf{0.52} & \underline{0.66} \\
MedGemma (27B)       & 0.39 & 0.46 & 0.42 & 0.65 & \textbf{0.52} & \underline{0.52} & \textbf{0.52} & 0.63 \\
Qwen-QwQ (32B)       & 0.35 & 0.49 & 0.41 & 0.66 & \underline{0.51} & 0.41 & 0.45 & 0.58 \\
II-Medical-8B        & \textbf{0.59} & 0.36 & 0.45 & 0.51 & 0.45 & 0.45 & 0.45 & 0.46 \\
\midrule
II-Medical-8B SFT (Ours) & \underline{0.54} & \textbf{0.58} & \textbf{0.56} & \textbf{0.74} & \underline{0.51} & 0.51 & \underline{0.51} & \textbf{0.69} \\
\bottomrule
\end{tabular}
\caption{Results on MedR-Bench and MedAction-300-Hard.}
\label{tab:results_compact}
\end{table*}

\subsection{Main Results}

As shown in Table~\ref{tab:results_compact}, among open-source models, we achieve the highest diagnosis accuracy, outperforming one of our teachers Baichuan-M3 
(235B). We also achieve consistently high test recommendation precision and recall, which demonstrates that high-quality multi-turn trajectory data drives larger 
gains than model scale alone.

GPT-5.4 is the strongest diagnosis model, establishing a ceiling with meaningful room for improvement. The base II-Medical-8B achieves high precision but very low recall, reflecting the ungrounded test ordering failure mode (Figure~\ref{fig:failure_mode}); MedAction training alleviates this issue substantially.

\subsection{Human Expert Validation}

To complement automatic evaluation, we conducted a blinded expert study on 100 randomly selected MedAction-300-Hard cases, resulting in 300 model trajectories (Base II-Medical-8B, OpenAI-o3-mini, and MedAction-SFT). An experienced licensed clinician independently evaluated each trajectory without access to the model identity or GPT-4o assessment. The evaluation covered diagnostic correctness, test appropriateness, safety, redundancy, and sequencing quality.

As shown in Table~\ref{tab:human_eval}, MedAction-SFT achieved the strongest overall clinical performance, substantially improving appropriate test ordering while reducing redundant examinations without sacrificing safety. These findings are consistent with the automatic evaluation results. Furthermore, GPT-4o showed strong agreement with the blinded clinician (Cohen's $\kappa=0.81$, 91\% raw agreement) while preserving the same ranking across all models, suggesting that GPT-4o provides a reliable approximation of human assessment.

\begin{table}[t]
\centering
\small
\begin{tabular}{lccc}
\toprule
Metric & II-Medical & o3-mini & MedAction-SFT \\
\midrule
Human diagnosis accuracy & 0.44 & 0.59 & \textbf{0.66}\\
Appropriate tests $\uparrow$ &39.9&48.3&\textbf{66.5}\\
Safe tests $\uparrow$&\textbf{89.6}&86.1&89.0\\
Redundant tests $\downarrow$&49.1&35.8&\textbf{18.4}\\
Good sequencing $\uparrow$&22&36&\textbf{49}\\
\midrule
GPT-4o diagnosis accuracy&0.48&0.62&\textbf{0.70}\\
\bottomrule
\end{tabular}
\caption{
Blinded clinician evaluation.
GPT-4o achieved Cohen's $\kappa=0.81$
(91\% agreement) with the clinician.
}
\label{tab:human_eval}
\end{table}

\begin{figure}
    \centering
    \includegraphics[width=1\linewidth]{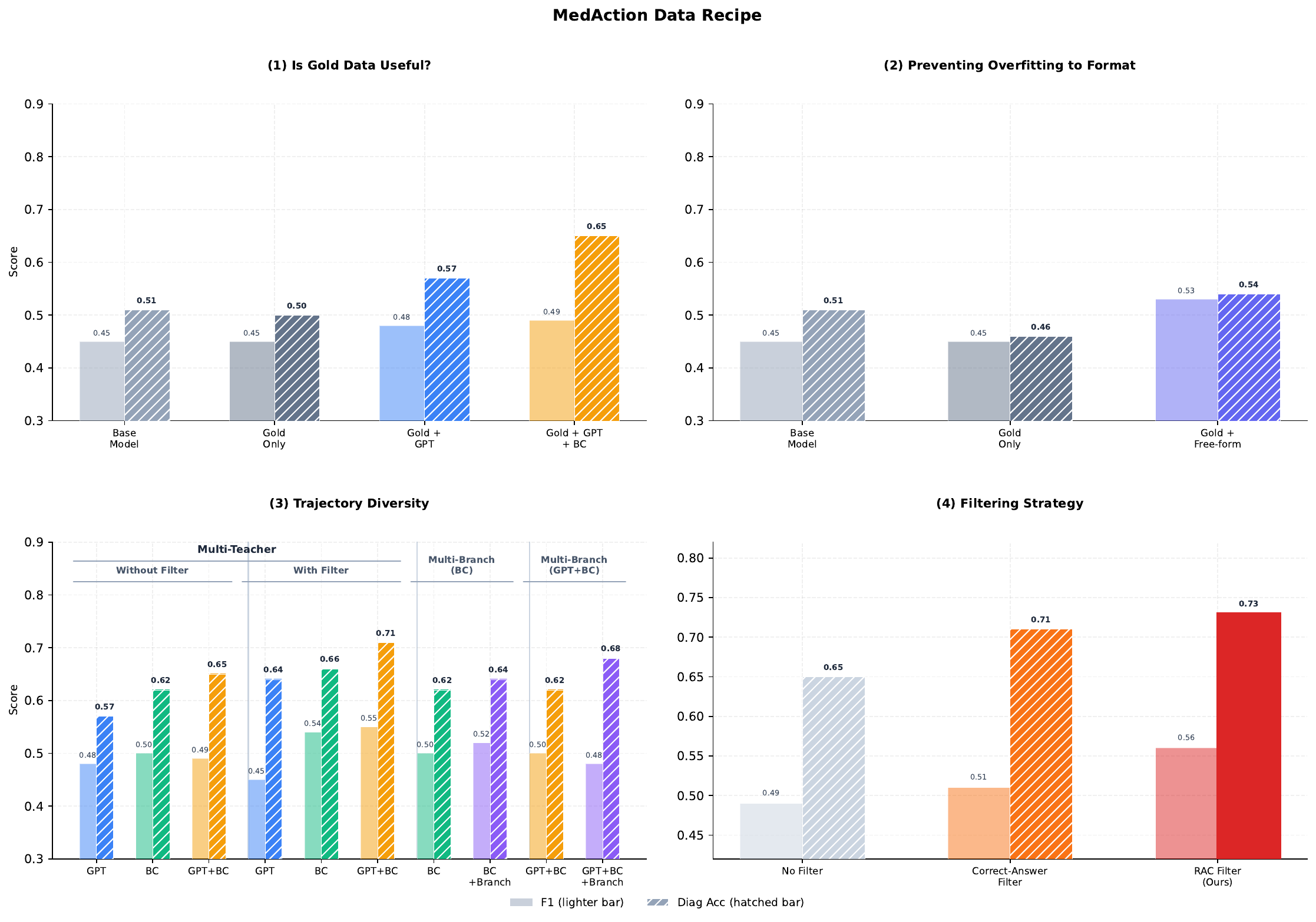}
    \caption{Ablation studies on trajectory data curation. Lighter bars: test recommendation F1; hatched bars: diagnosis accuracy.}
    \label{fig:ablation}
    \vspace{-15pt}
\end{figure}
\subsection{Analysis}
\label{sec:analysis}

A central question in MedAction is: \textit{how should we curate 
multi-turn diagnostic trajectories to teach LLM proper active diagnosis?} 
Hence, we explain via experiments four design choices in the pipeline. All results in the analysis section are on MedR-Bench. 

% ── RQ1 ──────────────────────────────────────────────────────────────────────
\paragraph{RQ1: Can "Gold" trajectories extracted directly from case reports 
serve as effective training data?}
An intuitive approach to constructing multi-turn training data is to 
extract established diagnostic trajectories from public medical case 
reports. We call these \textit{gold trajectories}: a strong LLM 
reconstructs the documented sequence of tests and findings from the 
raw case text, producing trajectories that are by construction 
ideal---every ordered test returns a result, every result is 
informative, and the trajectory converges to the correct diagnosis 
without detours. An alternative is to \textit{distill 
trajectories}: rather than reconstructing a documented path, we have 
teacher LLMs (GPT-5.4-mini and Baichuan-M3) interact with the 
clinical environment under genuine partial observability, where 
tests can be unavailable, results may be inconclusive, and 
diagnostic repivots are necessary. We compare four configurations: 
the base model (no SFT), gold-only, gold + GPT-distilled, and 
gold + GPT + Baichuan-distilled.
 
As shown in Figure~\ref{fig:ablation}-(1), training on gold 
trajectories alone does not improve diagnosis accuracy or test 
recommendation F1 over simply prompting the base model to perform structured diagnosis. We inspected model outputs and found out that the model learns the reasoning format of active differential diagnosis well but easily gets stuck when the 
environment deviates from a perfect path. 
\begin{tcolorbox}[colback=blue!5, colframe=blue!30, boxrule=0.5pt,
                  left=4pt, right=4pt, top=3pt, bottom=3pt]
\small\textbf{Takeaway:} Gold trajectories teach output format but 
produce weak reasoning. Environment-interactive distillation sampling is 
essential.
\end{tcolorbox}

% ── RQ2 ──────────────────────────────────────────────────────────────────────
\paragraph{RQ2: Does the output schema affect reasoning 
flexibility?}
RQ1 raises concerns about the model only learning the format but not the way to reason and act.
All trajectories in our dataset follow a fixed output schema at each turn: differential diagnosis list, pivot statement, and action 
selection. While this structure aids mimicking how clinicians reason, we found that the model's ``thinking mode'' behaviors, such 
as self-correction, reflection, and exploratory 
hypothesizing are slightly reduced. 
We address this by adding a small proportion (10\%) of \textit{free-form 
trajectories}: sequences generated without a prescribed output 
structure, where the model reasons unconstrained before committing 
to an action. As shown in Figure~\ref{fig:ablation}-(2), mixing in free-form 
data improves both test recommendation F1 and diagnosis accuracy 
over the gold-only baseline. The gain is most pronounced in F1, 
suggesting that free-form reasoning restores the model's ability 
to consider a broader set of tests rather than defaulting to 
schema-conditioned habits. 
\begin{tcolorbox}[colback=blue!5, colframe=blue!30, boxrule=0.5pt,
                  left=4pt, right=4pt, top=3pt, bottom=3pt]
\small\textbf{Takeaway:} A small proportion of free-form data 
mitigates format overfitting and rescues thinking-mode behaviors, yielding meaningful gains 
in test recommendation quality.
\end{tcolorbox}

% ── RQ3 ──────────────────────────────────────────────────────────────────────
\paragraph{RQ3: What sources of diversity matter most in rejection 
sampling?}

Diversity in training trajectories is particularly important for 
active diagnosis because the task is inherently \textit{multi-path}: 
unlike static QA where a single reasoning chain leads to the answer, 
the same initial presentation can support multiple valid diagnostic 
strategies---a clinician may screen broadly with routine labs before 
narrowing, or pursue a targeted confirmatory test immediately. 
Moreover, the action space is vast and open-vocabulary, meaning any 
single teacher LLM will cover only a narrow slice of valid test-ordering 
behavior. We examine two orthogonal sources of diversity that address 
different aspects of this challenge.

\textit{Multi-teacher diversity.} Different teacher LLMs exhibit 
different clinical reasoning styles---for instance, one may favor 
broad screening while another pursues targeted confirmatory tests. 
We compare trajectories generated by GPT-5.4-mini alone, 
Baichuan-M3 alone, and the combination, under two conditions: 
without any quality filtering, and with correctness filtering 
(retaining only trajectories that reach the ground-truth diagnosis).
\textit{Multi-teacher diversity} captures variation in clinical 
reasoning \textit{style}. Without filtering, combining teachers yields 
moderate gains over single-teacher baselines. With filtering, the 
same combination produces a substantially larger improvement, 
showing that multi-teacher diversity is most valuable \textit{after} 
low-quality trajectories are removed: unfiltered mixing introduces 
noise that dilutes the benefit of stylistic diversity.
 
\textit{Multi-branch diversity} captures variation in diagnostic 
\textit{decisions} from identical evidence states. We generate 
branched trajectories by launching new rollouts from intermediate 
turns of existing trajectories, producing multiple valid continuations 
from the same accumulated evidence. This directly mirrors the 
branching nature of clinical reasoning: given the same labs, two 
clinicians may legitimately order different next tests. Branching 
consistently improves performance in both single-teacher and 
multi-teacher configurations.
 
\begin{tcolorbox}[colback=blue!5, colframe=blue!30, boxrule=0.5pt,
                  left=4pt, right=4pt, top=3pt, bottom=3pt]
\small\textbf{Takeaway:} Multi-teacher diversity broadens reasoning style 
coverage and requires filtering to realize max benefits; multi-branch 
diversity also provides consistent gains.
\end{tcolorbox}

% ── RQ4 ──────────────────────────────────────────────────────────────────────

\paragraph{RQ4: Does knowledge-graph-grounded filtering outperform correctness-only filtering?}
From RQ3 analysis we identified filtering as an effective approach to generate high quality data and provide performance gains. We now compare \textit{what} to filter on. We evaluate three strategies applied to the same pool of GPT+BC trajectories with branching: (1)~\textit{no filtering}, which retains all sampled trajectories regardless of outcome; (2)~\textit{correctness filtering}, which retains only complete trajectories that reach the ground-truth diagnosis; and (3)~\textit{DTC+RAC filtering} (ours), which additionally retains convergent partially correct trajectories and removes turns where diagnostic belief changes were ungrounded in the evidence gathered (Section~\ref{sec:methodology}).
As shown in Figure~\ref{fig:ablation}-(4), unfiltered multi-teacher trajectories already achieve reasonable performance, but both filtering strategies yield meaningful additional gains. Correctness filtering improves diagnosis accuracy substantially but leaves F1 relatively low, because it retains only trajectories with correct endpoints while discarding all partially correct reasoning. Our filtering improves on both metrics, with the largest gain in F1. The mechanism is straightforward: by retaining convergent prefixes and pruning only ungrounded turns, RAC filtering recovers useful training signal that correctness filtering discards---particularly the test-ordering and belief-updating behavior in early and middle turns of otherwise imperfect trajectories.

\begin{figure}
    \centering
    \includegraphics[width=1\linewidth]{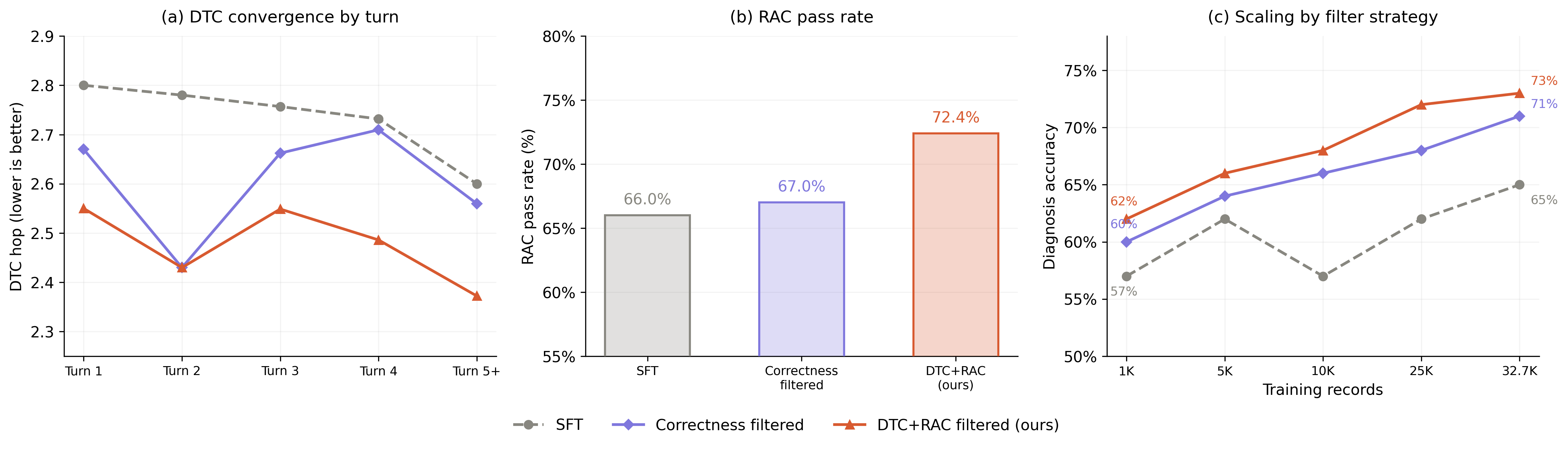}
    \caption{\textbf{Effect of trajectory filtering on diagnostic performance.} \textbf{\textit{(a)}} Average DTC for differently finetuned models at each turn. Lower values indicate closer alignment. \textbf{\textit{(b)}} Turn level pass rate under the RAC filtering. Higher is better. \textbf{\textit{(c)}} Scaling performance for each filtering strategy.}
    \label{fig:post_train}
    \vspace{-15pt}
\end{figure}

To understand \textit{how} this translates into improved model behavior, we evaluate 100 held-out cases across the three training configurations and measure trajectory-level diagnostic quality (Figure~\ref{fig:post_train}). Panel~(a) tracks the DTC hop distance between each model's top hypothesis and the ground-truth diagnosis at each turn. The SFT baseline starts with the highest DTC and converges slowly, while correctness-filtered and DTC+RAC-filtered models both start lower. The DTC+RAC model reaches the lowest final DTC, confirming that it learns effective diagnostic reasoning over longer horizons rather than committing early to a wrong answer. Panel~(b) applies RAC filtering to all model outputs, and the pass rate measures the fraction of turns where belief updates were grounded in the evidence gathered. The DTC+RAC-filtered model achieves the highest pass rate, directly addressing the reasoning-action mismatch. Panel~(c) examines data efficiency: we train models at five data scales (1K--32K records) under all three filtering strategies. DTC+RAC filtering dominates at every scale, with the largest advantage at low data budgets. Unfiltered scaling is non-monotonic, dipping at 10K before recovering, while DTC+RAC scaling increases monotonically, confirming that filtering removes harmful training signal rather than merely selecting easier cases.
\begin{tcolorbox}[colback=blue!5, colframe=blue!30, boxrule=0.5pt, left=4pt, right=4pt, top=3pt, bottom=3pt]
\small\textbf{Takeaway:} DTC+RAC filtering outperforms correctness-only filtering by recovering useful intermediate reasoning from partially correct trajectories and reducing ungrounded turn-level actions. Models trained on DTC+RAC-filtered data exhibit stronger diagnostic convergence, reduction in ungrounded actions, and more data-efficient scaling.
\end{tcolorbox}

\subsection{Extension to Reinforcement Learning}
\label{sec:rl_extension}

\textit{(Added in arXiv v2; not part of the peer-reviewed COLM 2026 version.)}
MedAction's environments and metrics also serve directly as building blocks for reinforcement learning. Starting from a turn-level SFT cold start on DTC/RAC-filtered MedAction trajectories, we train Qwen3.5-27B~\citep{qwen35} with GRPO~\citep{shao2024deepseekmath} using a binary diagnosis reward and a DTC-gated test reward, which credits a documented test only when the subsequent belief update does not move the top hypothesis farther from the ground truth on PrimeKG. The resulting 27B agent reaches 0.77 and 0.73 diagnostic accuracy on MedR-Bench and MedAction-300-Hard. We further find that (i) the cold-start gain stems mainly from DTC/RAC curation, which turn-level training makes possible, (ii) the benefit of RL grows with model size and differs between general-domain and medical backbones, and (iii) compared with a Jaccard test-overlap reward, the DTC-gated reward reduces the share of tests followed by divergent belief updates from 15.2\% to 4.6\%. Because these experiments use a different agent interface, their numbers are not directly comparable to Table~\ref{tab:results_compact}; full details are given in Appendix~\ref{app:rl}.

\section{Conclusion and Limitations}
We presented MedAction, a data construction framework for improving LLM performance on active diagnosis. Our analysis identifies three systematic failure modes in current LLMs, and our environment-interactive trajectory distillation --- grounded by two new metrics (DTC and RAC) --- directly addresses them. Our 8B model fine-tuned on MedAction-32K achieves state-of-the-art open-source performance, providing strong foundations for future work. In Appendix~\ref{app:rl}, we further show that MedAction's environments and metrics directly support reinforcement learning: an SFT-then-GRPO recipe with a DTC-gated test reward yields a 27B agent with higher diagnostic accuracy and fewer belief-regressing test orders. Remaining challenges lie mainly in the environment and evaluation: case reports document only a subset of clinically reasonable tests, so building more complete, clinically grounded environments (e.g., EHR-derived or clinician-authored) and conducting broader clinical validation are important next steps toward real-world applicability.

\section*{Acknowledgments}
We thank Bohao Wang, Pengfei Hu, Xiaoyu Qiu, Yichen Guo, David Restrepo, and Luis Filipe Nakayama for their contributions to the reinforcement learning extension presented in Appendix~\ref{app:rl}, which was added in arXiv v2.

\bibliography{colm2026_conference}

@article{singhal2023large,
  title={Large language models encode clinical knowledge},
  author={Singhal, Karan and Azizi, Shekoofeh and Tu, Tao and Mahdavi, S Sara and Wei, Jason and Chung, Hyung Won and Scales, Nathan and Tanwani, Ajay and Cole-Lewis, Heather and Pfohl, Stephen and others},
  journal={Nature},
  volume={620},
  number={7972},
  pages={172--180},
  year={2023},
  publisher={Nature Publishing Group UK London}
}

@article{jin2021disease,
  title={What disease does this patient have? a large-scale open domain question answering dataset from medical exams},
  author={Jin, Di and Pan, Eileen and Oufattole, Nassim and Weng, Wei-Hung and Fang, Hanyi and Szolovits, Peter},
  journal={Applied Sciences},
  volume={11},
  number={14},
  pages={6421},
  year={2021},
  publisher={MDPI}
}

@article{qiu2025quantifying,
  title={Quantifying the reasoning abilities of LLMs on clinical cases},
  author={Qiu, Pengcheng and Wu, Chaoyi and Liu, Shuyu and Fan, Yanjie and Zhao, Weike and Chen, Zhuoxia and Gu, Hongfei and Peng, Chuanjin and Zhang, Ya and Wang, Yanfeng and others},
  journal={Nature Communications},
  volume={16},
  number={1},
  pages={9799},
  year={2025},
  publisher={Nature Publishing Group UK London}
}

@inproceedings{pal2022medmcqa,
  title={Medmcqa: A large-scale multi-subject multi-choice dataset for medical domain question answering},
  author={Pal, Ankit and Umapathi, Logesh Kumar and Sankarasubbu, Malaikannan},
  booktitle={Conference on health, inference, and learning},
  pages={248--260},
  year={2022},
  organization={PMLR}
}

@inproceedings{jin2019pubmedqa,
  title={Pubmedqa: A dataset for biomedical research question answering},
  author={Jin, Qiao and Dhingra, Bhuwan and Liu, Zhengping and Cohen, William and Lu, Xinghua},
  booktitle={Proceedings of the 2019 conference on empirical methods in natural language processing and the 9th international joint conference on natural language processing (EMNLP-IJCNLP)},
  pages={2567--2577},
  year={2019}
}

@article{li2024mediq,
  title={Mediq: Question-asking llms and a benchmark for reliable interactive clinical reasoning},
  author={Li, Shuyue S and Balachandran, Vidhisha and Feng, Shangbin and Ilgen, Jonathan S and Pierson, Emma and Koh, Pang W and Tsvetkov, Yulia},
  journal={Advances in Neural Information Processing Systems},
  volume={37},
  pages={28858--28888},
  year={2024}
}

@article{wu2024medjourney,
  title={Medjourney: Benchmark and evaluation of large language models over patient clinical journey},
  author={Wu, Xian and Zhao, Yutian and Zhang, Yunyan and Wu, Jiageng and Zhu, Zhihong and Zhang, Yingying and Ouyang, Yi and Zhang, Ziheng and Wang, Huimin and Lin, Zhenxi and others},
  journal={Advances in Neural Information Processing Systems},
  volume={37},
  pages={87621--87646},
  year={2024}
}

@inproceedings{guha2026openthoughts,
title={OpenThoughts: Data Recipes for Reasoning Models},
author={Etash Kumar Guha and Ryan Marten and Sedrick Keh and Negin Raoof and Georgios Smyrnis and Hritik Bansal and Marianna Nezhurina and Jean Mercat and Trung Vu and Zayne Rea Sprague and Ashima Suvarna and Benjamin Feuer and Leon Liangyu Chen and Zaid Khan and Eric Frankel and Sachin Grover and Caroline Choi and Niklas Muennighoff and Shiye Su and Wanjia Zhao and John Yang and Shreyas Pimpalgaonkar and Kartik sharma and Charlie Cheng-Jie Ji and Yichuan Deng and Sarah M Pratt and Vivek Ramanujan and Jon Saad-Falcon and Stutee Acharya and Jeffrey Li and Achal Dave and Alon Albalak and Kushal Arora and Blake Wulfe and Chinmay Hegde and Greg Durrett and Sewoong Oh and Mohit Bansal and Saadia Gabriel and Aditya Grover and Kai-Wei Chang and Vaishaal Shankar and Aaron Gokaslan and Mike A Merrill and Tatsunori Hashimoto and Yejin Choi and Jenia Jitsev and Reinhard Heckel and Maheswaran Sathiamoorthy and Alex Dimakis and Ludwig Schmidt},
booktitle={The Fourteenth International Conference on Learning Representations},
year={2026},
url={https://openreview.net/forum?id=7xjoTuaNmN}
}

@inproceedings{muennighoff2025s1,
  title={s1: Simple test-time scaling},
  author={Muennighoff, Niklas and Yang, Zitong and Shi, Weijia and Li, Xiang Lisa and Fei-Fei, Li and Hajishirzi, Hannaneh and Zettlemoyer, Luke and Liang, Percy and Cand{\`e}s, Emmanuel and Hashimoto, Tatsunori B},
  booktitle={Proceedings of the 2025 Conference on Empirical Methods in Natural Language Processing},
  pages={20286--20332},
  year={2025}
}

@article{wu2025medreason,
  title={Medreason: Eliciting factual medical reasoning steps in llms via knowledge graphs},
  author={Wu, Juncheng and Deng, Wenlong and Li, Xingxuan and Liu, Sheng and Mi, Taomian and Peng, Yifan and Xu, Ziyang and Liu, Yi and Cho, Hyunjin and Choi, Chang-In and others},
  journal={arXiv preprint arXiv:2504.00993},
  year={2025}
}

@article{ossowski2025octomed,
  title={OctoMed: Data Recipes for State-of-the-Art Multimodal Medical Reasoning},
  author={Ossowski, Timothy and Zhang, Sheng and Liu, Qianchu and Qin, Guanghui and Tan, Reuben and Naumann, Tristan and Hu, Junjie and Poon, Hoifung},
  journal={arXiv preprint arXiv:2511.23269},
  year={2025}
}

@inproceedings{chen-etal-2025-towards-medical,
    title = "Towards Medical Complex Reasoning with {LLM}s through Medical Verifiable Problems",
    author = "Chen, Junying  and
      Cai, Zhenyang  and
      Ji, Ke  and
      Wang, Xidong  and
      Liu, Wanlong  and
      Wang, Rongsheng  and
      Wang, Benyou",
    editor = "Che, Wanxiang  and
      Nabende, Joyce  and
      Shutova, Ekaterina  and
      Pilehvar, Mohammad Taher",
    booktitle = "Findings of the Association for Computational Linguistics: ACL 2025",
    month = jul,
    year = "2025",
    address = "Vienna, Austria",
    publisher = "Association for Computational Linguistics",
    url = "https://aclanthology.org/2025.findings-acl.751/",
    doi = "10.18653/v1/2025.findings-acl.751",
    pages = "14552--14573",
    ISBN = "979-8-89176-256-5"
}

@article{huang2025m1,
  title={m1: Unleash the potential of test-time scaling for medical reasoning with large language models},
  author={Huang, Xiaoke and Wu, Juncheng and Liu, Hui and Tang, Xianfeng and Zhou, Yuyin},
  journal={arXiv preprint arXiv:2504.00869},
  year={2025}
}

@techreport{openai2025o3mini,
  title   = {OpenAI o3-mini System Card},
  author  = {OpenAI},
  year    = {2025},
  url     = {https://cdn.openai.com/o3-mini-system-card-feb10.pdf},
  institution  = {OpenAI}
}

@misc{nlm2003pmcoa,
  title        = {PMC Open Access Subset [Internet]},
  author       = {{National Library of Medicine}},
  year         = {2003},
  howpublished = {\url{https://pmc.ncbi.nlm.nih.gov/tools/openftlist/}},
  note         = {Bethesda (MD)}
}

@article{chandak2023building,
  title={Building a knowledge graph to enable precision medicine},
  author={Chandak, Payal and Huang, Kexin and Zitnik, Marinka},
  journal={Scientific Data},
  volume={10},
  number={1},
  pages={67},
  url={https://doi.org/10.1038/s41597-023-01960-3},
  year={2023},
  publisher={Nature Publishing Group}
}

@article{johnson2024unified,
  title={ClinVec: Unified Embeddings of Clinical Codes Enable Knowledge-Grounded AI in Medicine},
  author={Johnson, Ruth and Gottlieb, Uri and Shaham, Galit and Eisen, Lihi and Waxman, Jacob and Devons-Sberro, Stav and Ginder, Curtis R. and Hong, Peter and Sayeed, Raheel and Reis, Ben Y. and Balicer, Ran D. and Dagan, Noa and Zitnik, Marinka},
  journal={medrxiv},
  url={https://www.medrxiv.org/content/10.1101/2024.12.03.24318322},
  year={2024}
}

@inproceedings{wu2025collabllm,
title={Collab{LLM}: From Passive Responders to Active Collaborators},
author={Shirley Wu and Michel Galley and Baolin Peng and Hao Cheng and Gavin Li and Yao Dou and Weixin Cai and James Zou and Jure Leskovec and Jianfeng Gao},
booktitle={Forty-second International Conference on Machine Learning},
year={2025},
url={https://openreview.net/forum?id=DmH4HHVb3y}
}

@inproceedings{liu2024agentbench,
title={AgentBench: Evaluating {LLM}s as Agents},
author={Xiao Liu and Hao Yu and Hanchen Zhang and Yifan Xu and Xuanyu Lei and Hanyu Lai and Yu Gu and Hangliang Ding and Kaiwen Men and Kejuan Yang and Shudan Zhang and Xiang Deng and Aohan Zeng and Zhengxiao Du and Chenhui Zhang and Sheng Shen and Tianjun Zhang and Yu Su and Huan Sun and Minlie Huang and Yuxiao Dong and Jie Tang},
booktitle={The Twelfth International Conference on Learning Representations},
year={2024},
url={https://openreview.net/forum?id=zAdUB0aCTQ}
}

@inproceedings{yao2025taubench,
title={\{\${\textbackslash}tau\$\}-bench: A Benchmark for {\textbackslash}underline\{T\}ool-{\textbackslash}underline\{A\}gent-{\textbackslash}underline\{U\}ser Interaction in Real-World Domains},
author={Shunyu Yao and Noah Shinn and Pedram Razavi and Karthik R Narasimhan},
booktitle={The Thirteenth International Conference on Learning Representations},
year={2025},
url={https://openreview.net/forum?id=roNSXZpUDN}
}

@article{jin2023medcpt,
  title={Medcpt: Contrastive pre-trained transformers with large-scale pubmed search logs for zero-shot biomedical information retrieval},
  author={Jin, Qiao and Kim, Won and Chen, Qingyu and Comeau, Donald C and Yeganova, Lana and Wilbur, W John and Lu, Zhiyong},
  journal={Bioinformatics},
  volume={39},
  number={11},
  pages={btad651},
  year={2023},
  publisher={Oxford University Press}
}

@article{qwen3technicalreport,
  title={Qwen3 technical report},
  author={Yang, An and Li, Anfeng and Yang, Baosong and Zhang, Beichen and Hui, Binyuan and Zheng, Bo and Yu, Bowen and Gao, Chang and Huang, Chengen and Lv, Chenxu and others},
  journal={arXiv preprint arXiv:2505.09388},
  year={2025}
}

@misc{2025II-Medical-8B,
      title={II-Medical-8B: Medical Reasoning Model}, 
      author={Intelligent Internet},
      year={2025}
}

@article{hurst2024gpt,
  title={Gpt-4o system card},
  author={Hurst, Aaron and Lerer, Adam and Goucher, Adam P and Perelman, Adam and Ramesh, Aditya and Clark, Aidan and Ostrow, AJ and Welihinda, Akila and Hayes, Alan and Radford, Alec and others},
  journal={arXiv preprint arXiv:2410.21276},
  year={2024}
}

@misc{openaigpt54mini,
  title        = {Introducing GPT-5.4 mini and nano},
  author       = {{OpenAI}},
  year         = {2026},
  howpublished = {\url{https://openai.com/index/introducing-gpt-5-4-mini-and-nano/}},
}

@article{dou2026baichuan,
  title={Baichuan-M3: Modeling Clinical Inquiry for Reliable Medical Decision-Making},
  author={Dou, Chengfeng and Yang, Fan and Li, Fei and Jia, Jiyuan and Ju, Qiang and Wang, Shuai and Li, Tianpeng and Zeng, Xiangrong and Zhou, Yijie and Zhang, Hongda and others},
  journal={arXiv preprint arXiv:2602.06570},
  year={2026}
}

@misc{qwq32b,
    title = {QwQ-32B: Embracing the Power of Reinforcement Learning},
    url = {https://qwenlm.github.io/blog/qwq-32b/},
    author = {Qwen Team},
    month = {March},
    year = {2025}
}

@article{qwen2.5,
      title={Qwen2.5 Technical Report}, 
      author={An Yang and Baosong Yang and Beichen Zhang and Binyuan Hui and Bo Zheng and Bowen Yu and Chengyuan Li and Dayiheng Liu and Fei Huang and Haoran Wei and Huan Lin and Jian Yang and Jianhong Tu and Jianwei Zhang and Jianxin Yang and Jiaxi Yang and Jingren Zhou and Junyang Lin and Kai Dang and Keming Lu and Keqin Bao and Kexin Yang and Le Yu and Mei Li and Mingfeng Xue and Pei Zhang and Qin Zhu and Rui Men and Runji Lin and Tianhao Li and Tianyi Tang and Tingyu Xia and Xingzhang Ren and Xuancheng Ren and Yang Fan and Yang Su and Yichang Zhang and Yu Wan and Yuqiong Liu and Zeyu Cui and Zhenru Zhang and Zihan Qiu},
      journal={arXiv preprint arXiv:2412.15115},
      year={2024}
}

@article{comanici2025gemini,
  title={Gemini 2.5: Pushing the frontier with advanced reasoning, multimodality, long context, and next generation agentic capabilities},
  author={Comanici, Gheorghe and Bieber, Eric and Schaekermann, Mike and Pasupat, Ice and Sachdeva, Noveen and Dhillon, Inderjit and Blistein, Marcel and Ram, Ori and Zhang, Dan and Rosen, Evan and others},
  journal={arXiv preprint arXiv:2507.06261},
  year={2025}
}

@techreport{openai2026gpt54thinking,
  title        = {{GPT-5.4 Thinking System Card}},
  author       = {{OpenAI}},
  year         = {2026},
  month        = mar,
  url          = {https://openai.com/index/gpt-5-4-thinking-system-card/},
  institution  = {OpenAI}
}

@article{wang2025baichuan,
  title={Baichuan-m1: Pushing the medical capability of large language models},
  author={Wang, Bingning and Zhao, Haizhou and Zhou, Huozhi and Song, Liang and Xu, Mingyu and Cheng, Wei and Zeng, Xiangrong and Zhang, Yupeng and Huo, Yuqi and Wang, Zecheng and others},
  journal={arXiv preprint arXiv:2502.12671},
  year={2025}
}

@article{sellergren2025medgemma,
  title={Medgemma technical report},
  author={Sellergren, Andrew and Kazemzadeh, Sahar and Jaroensri, Tiam and Kiraly, Atilla and Traverse, Madeleine and Kohlberger, Timo and Xu, Shawn and Jamil, Fayaz and Hughes, C{\'\i}an and Lau, Charles and others},
  journal={arXiv preprint arXiv:2507.05201},
  year={2025}
}

@inproceedings{chen-etal-2025-cod,
    title = "{C}o{D}, Towards an Interpretable Medical Agent using Chain of Diagnosis",
    author = "Chen, Junying  and
      Gui, Chi  and
      Gao, Anningzhe  and
      Ji, Ke  and
      Wang, Xidong  and
      Wan, Xiang  and
      Wang, Benyou",
    editor = "Che, Wanxiang  and
      Nabende, Joyce  and
      Shutova, Ekaterina  and
      Pilehvar, Mohammad Taher",
    booktitle = "Findings of the Association for Computational Linguistics: ACL 2025",
    month = jul,
    year = "2025",
    address = "Vienna, Austria",
    publisher = "Association for Computational Linguistics",
    url = "https://aclanthology.org/2025.findings-acl.740/",
    doi = "10.18653/v1/2025.findings-acl.740",
    pages = "14345--14368",
    ISBN = "979-8-89176-256-5"
}

@article{schmidgall2024agentclinic,
  title={Agentclinic: a multimodal agent benchmark to evaluate ai in simulated clinical environments},
  author={Schmidgall, Samuel and Ziaei, Rojin and Harris, Carl and Reis, Eduardo and Jopling, Jeffrey and Moor, Michael},
  journal={arXiv preprint arXiv:2405.07960},
  year={2024}
}

@inproceedings{lai2026doctorr,
title={Doctor-R1: Mastering Clinical Inquiry with Experiential Agentic Reinforcement Learning},
author={Yunghwei Lai and Kaiming Liu and Ziyue Wang and Weizhi Ma and Yang Liu},
booktitle={The Fourteenth International Conference on Learning Representations},
year={2026},
url={https://openreview.net/forum?id=vQGHTyL0Jw}
}

@article{tu2025towards,
  title={Towards conversational diagnostic artificial intelligence},
  author={Tu, Tao and Schaekermann, Mike and Palepu, Anil and Saab, Khaled and Freyberg, Jan and Tanno, Ryutaro and Wang, Amy and Li, Brenna and Amin, Mohamed and Cheng, Yong and others},
  journal={Nature},
  volume={642},
  number={8067},
  pages={442--450},
  year={2025},
  publisher={Nature Publishing Group UK London}
}

@article{zhang2024ultramedical,
  title={Ultramedical: Building specialized generalists in biomedicine},
  author={Zhang, Kaiyan and Zeng, Sihang and Hua, Ermo and Ding, Ning and Chen, Zhang-Ren and Ma, Zhiyuan and Li, Haoxin and Cui, Ganqu and Qi, Biqing and Zhu, Xuekai and others},
  journal={Advances in Neural Information Processing Systems},
  volume={37},
  pages={26045--26081},
  year={2024}
}

@inproceedings{
liu2025multiturn,
title={Multi-Turn {LLM} Systems for Diagnostic Decision-Making: Considerations, Biases, and Challenges},
author={Benjamin Liu and Sejong Kim and Drona Thoka and Varun Puttagunta and Kaylin Sheng and Mark Li and Kiran Nijjer and Adnan Ahmed and Thi Uyen Hanh Le and Sai Chidvilas Gudiboina and Ali Ugur and Kevin Zhu},
booktitle={First Workshop on Multi-Turn Interactions in Large Language Models},
year={2025},
url={https://openreview.net/forum?id=It7AgR3A9H}
}

@article{shao2024deepseekmath,
  title={Deepseekmath: Pushing the limits of mathematical reasoning in open language models},
  author={Shao, Zhihong and Wang, Peiyi and Zhu, Qihao and Xu, Runxin and Song, Junxiao and Bi, Xiao and Zhang, Haowei and Zhang, Mingchuan and Li, YK and Wu, Yang and others},
  journal={arXiv preprint arXiv:2402.03300},
  year={2024}
}

@misc{qwen35,
  title  = {{Qwen3.5}: Towards Native Multimodal Agents},
  author = {{Qwen Team}},
  month  = {February},
  year   = {2026},
  url    = {https://qwen.ai/blog?id=qwen3.5}
}

@article{gao2026medexagent,
  title={MedExAgent: Training LLM Agents to Ask, Examine, and Diagnose in Noisy Clinical Environments},
  author={Gao, Yicheng and Zhou, Xiaolin and Li, Yahan and Zhao, Yue and Liu, Ruishan},
  journal={arXiv preprint arXiv:2605.07058},
  year={2026}
}

@article{chen2026meissa,
  title={Meissa: Multi-modal medical agentic intelligence},
  author={Chen, Yixiong and Bai, Xinyi and Pan, Yue and Zhou, Zongwei and Yuille, Alan},
  journal={arXiv preprint arXiv:2603.09018},
  year={2026}
}

@misc{cms2025feeschedule,
  title        = {{CY 2025 Fee Schedules}},
  author       = {{Centers for Medicare \& Medicaid Services}},
  year         = {2025},
  howpublished = {\url{https://www.cms.gov/medicare/payment/fee-schedules}}
}

@inproceedings{kwon2023efficient,
  title={Efficient memory management for large language model serving with pagedattention},
  author={Kwon, Woosuk and Li, Zhuohan and Zhuang, Siyuan and Sheng, Ying and Zheng, Lianmin and Yu, Cody Hao and Gonzalez, Joseph and Zhang, Hao and Stoica, Ion},
  booktitle={Proceedings of the 29th symposium on operating systems principles},
  pages={611--626},
  year={2023}
}
\bibliographystyle{colm2026_conference}
\clearpage
\appendix

% % You may include other additional sections here.

% \appendix

% 你的 failure mode 内容

\section{Prompt Templates}
\label{app:prompt_template}

\begin{promptbox}{System prompt for the diagnostic agent.}
You are an experienced clinical physician performing a structured diagnostic workup.
\end{promptbox}

\begin{promptbox}{Initial-turn prompt for structured diagnostic reasoning.}
You are seeing this patient for the first time. Analyze the case and respond using EXACTLY the section headers below in EXACTLY this order. Do not skip or rename any section.

Patient Case Summary:
{case}

---

### Chain of Thought:
[Think step by step. Label each step <step 1>, <step 2>, etc.]

### DDx List:
[Ranked differentials. Be specific.]
1. Diagnosis name - one sentence reasoning
2. Diagnosis name - one sentence reasoning
3. Diagnosis name - one sentence reasoning

### Pivot:
[One paragraph: the single most important diagnostic question right now and why.]

### Primary Actions:
[Most targeted, high-yield tests to resolve top differentials THIS turn. Specific test names only.]
1. test name - purpose
2. test name - purpose

### Additional Information Required:
[Any broader workup beyond Primary Actions: routine labs, imaging, history.
If a test comes back normal or shows no significant findings, do NOT order
any further tests in the same category - move to a completely different
diagnostic category or test a different aspect of the presentation.
Write "Not required." only if you are fully confident in the final diagnosis.]
1. category: specific test or data needed

### Diagnostic Status:
[Write EXACTLY one word: either DONE or CONTINUE.
Write DONE only if you have sufficient information to make a definitive final diagnosis right now.
Write CONTINUE if you need more test results before concluding.]

### Conclusion:
[Your single best current diagnosis. Be specific. One sentence only.]
\end{promptbox}
\clearpage 

\begin{promptbox}{Follow-up-turn prompt with cumulative test history and new results.}
Patient Case Summary:
{case}

--- GLOBAL STATUS (cumulative across all turns) ---

All Tests Ordered So Far:
{done_tests_block}

Tests Confirmed UNAVAILABLE (do NOT re-order these):
{unavailable_tests_block}
NOTE: When key diagnostic tests are UNAVAILABLE, this is a signal that you cannot confirm your current leading hypothesis through the expected pathway. You should treat UNAVAILABLE tests as a reason to PIVOT your diagnostic direction --- consider alternative diagnoses that can be confirmed or excluded using different, available tests rather than persisting with the same differential.

--- RECENT DIAGNOSTIC HISTORY (last {window_size} turn(s)) ---

{recent_turns_block}

--- NEW RESULTS THIS TURN ---

{oracle_results}

---

Respond using EXACTLY the section headers below in EXACTLY this order.

### Chain of Thought:
[Step-by-step reasoning.
CRITICAL RULES:
- If key tests are UNAVAILABLE, do NOT simply wait or repeat the same differential.]

### DDx List:
[Updated differentials. Reflect that Normal results have excluded certain possibilities and that UNAVAILABLE tests may mean your leading hypothesis cannot be verified --- consider alternative diagnoses.]

### Pivot:
[Explain how Normal results and/or UNAVAILABLE tests shifted your diagnostic direction and what different path you are pursuing now.]

### Primary Actions:
[New targeted tests ONLY.
RULES:
1. If a test is listed under "Tests Confirmed UNAVAILABLE" above, you are FORBIDDEN from re-ordering it.
2. If a test is listed under "All Tests Ordered So Far", do NOT re-order it.
3. You must either:
   a) Order a different, alternative diagnostic test that IS available.
   b) Move to a final "Conclusion" if no other available tests can help.
4. DO NOT use words like "Await", "Wait for", or "Review previous tests".]

### Additional Information Required:
[Any broader workup. Write "Not required." if you are shifting to a final conclusion.]

### Diagnostic Status:
[Write EXACTLY one word: either DONE or CONTINUE.
Write DONE only if you have sufficient information right now to make a definitive final diagnosis.
Write CONTINUE if further workup is still needed.]

### Conclusion:
[Single best current diagnosis given all available info.]
\end{promptbox}

\begin{promptbox}{Oracle prompt for returning ancillary test findings}
You are a hospital administrator providing status updates on diagnostic requests.

Guidelines:
1. If the requested test is present in the Ancillary Test Results below, provide the specific findings.
2. If the test is NOT in the data, respond with:
   "[test name]: This test is currently UNAVAILABLE due to equipment maintenance or lack of specialized personnel. You must proceed with clinical diagnosis or alternative available testing."

Patient Case
{full_case_summary}

Ancillary Test Results
{anc}
\end{promptbox}

\begin{promptbox}{Prompt for extracting ordered tests into a JSON array.}
You are a medical text organizer.
Given a physician's full diagnostic response, extract ALL diagnostic tests or examinations they have requested or ordered into a flat JSON array.

Rules:
- Extract ONLY test/examination names, not purposes or reasoning.
- Split compound entries: "CBC, CMP, LFTs" into three separate items.
- Look in all sections: Primary Actions, Additional Information Required, or anywhere tests are mentioned.
- If no tests are requested, return [].
- Be specific: "serum troponin I" not just "troponin".

Return ONLY a valid JSON array, no markdown, no extra text:
["test name 1", "test name 2", ...]
\end{promptbox}
\clearpage 

\begin{promptbox}{Prompt for matching predicted tests against ground-truth tests.}
You are a medical test matching expert.

Given PREDICTED tests and GROUND TRUTH (GT) tests, determine which GT items are covered.

RULES:
1. Name equivalence: "CBC" == "Complete Blood Count", "CXR" == "Chest X-ray"
2. Parent covers children: "MRI Brain" covers "MRI Brain T1", "MRI Brain FLAIR"
3. Compound GT items (comma/slash separated): treat as ONE unit.
4. One predicted test can cover multiple GT items. One GT item counted once only.
5. Do NOT mark covered if overlap is minor.

Return ONLY valid JSON:
{
  "gt_covered":   ["..."],
  "gt_uncovered": ["..."],
  "pred_used":    ["..."],
  "pred_unused":  ["..."]
}
\end{promptbox}

\begin{promptbox}{Prompt for evaluating diagnostic equivalence.}
You are a medical evaluation assistant.
Determine if Predicted Diagnosis and Ground Truth Diagnosis are clinically equivalent.
Return ONLY valid JSON: {"match": true, "reasoning": "..."}
\end{promptbox}

\paragraph{Prompts for the RL agent interface (Appendix~\ref{app:rl}).}
\label{app:rl_prompt}
The RL experiments in Appendix~\ref{app:rl} use the following tool-call interface instead of the structured prompts above: the agent orders at most one test per turn and reports a concise diagnosis and status every turn.

\begin{promptbox}{System prompt for the RL diagnostic agent (Appendix~\ref{app:rl}).}
You are an expert clinical physician working through a diagnostic case.

You will be given a patient's initial presentation. You will reason through the case across multiple turns (you have at most **8 turns**), ordering tests and refining your diagnosis. **EVERY turn**, you must output ALL THREE of the following blocks, in this exact order:

## 1. Tool call
If you want to order a test this turn (true unless you decide that evidence are enough), include exactly one tool call:

<tool_call>
{"name": "order_test", "arguments": {"test_name": "<canonical test name>"}}
</tool_call>

If you have enough information and want to stop ordering tests, omit the tool call.

## 2. Current best diagnosis (REQUIRED every turn)
Your best diagnosis given the evidence so far. Use this EXACT XML format on a single line:

<diagnosis>YOUR CURRENT BEST DIAGNOSIS</diagnosis>

The diagnosis must be CONCISE -- disease name + key qualifier only. Good examples:

- <diagnosis>Acute appendicitis</diagnosis>

- <diagnosis>Hashimoto thyroiditis</diagnosis>

- <diagnosis>Orbital cellulitis secondary to acute sinusitis</diagnosis>

Do NOT use markdown code fences, headers like "### Final Diagnosis", narrative phrases like "The patient is diagnosed with...", or markdown bold like **diagnosis**.

## 3. Status (REQUIRED every turn)
Signal whether you want to continue or are done:

<status>continue</status>   -- you want to order more tests and refine the diagnosis next turn
<status>done</status>       -- you are confident and want to commit your <diagnosis> as the final answer

Use <status>done</status> when you have enough evidence. Use <status>continue</status> if you need more tests.

## Turn budget
You have at most **8 turns** to investigate. Plan to commit: emit <status>done</status> as soon as the evidence is sufficient, and ALWAYS before you reach the 8-turn limit. A focused workup is usually 2-4 tests; ordering more rarely changes the diagnosis. If you reach the limit without committing <status>done</status>, your most recent <diagnosis> is taken as your final answer -- so make sure it is already your best one.

## Tool: order_test.
The tool returns the result, or "UNAVAILABLE" if the test is not part of this case's workup. Use canonical clinical names -- concrete anatomy + modality, no abbreviations alone.

## Example turn (with tool call):

<tool_call>
{"name": "order_test", "arguments": {"test_name": "<canonical test name>"}}
</tool_call>

Brief reasoning that ties the patient's findings to the test you ordered and to your current working diagnosis.

<diagnosis><concise disease name and qualifier></diagnosis>
<status>continue</status>

## Example turn (final, no more tests):

Brief reasoning explaining why the accumulated evidence is sufficient and which single diagnosis it points to.

<diagnosis><concise disease name and qualifier></diagnosis>
<status>done</status>
\end{promptbox}

\begin{promptbox}{User prompt for the RL diagnostic agent (Appendix~\ref{app:rl}).}
## Patient Presentation

{initial_obs}

Begin your diagnostic workup. Remember: every turn must include <diagnosis>...</diagnosis> and <status>...</status>.
\end{promptbox}

\clearpage

\section{Dataset Composition and Test Distribution}
\label{app:dataset_comp}

\begin{figure}[t]
    \centering
    \includegraphics[width=0.8\linewidth]{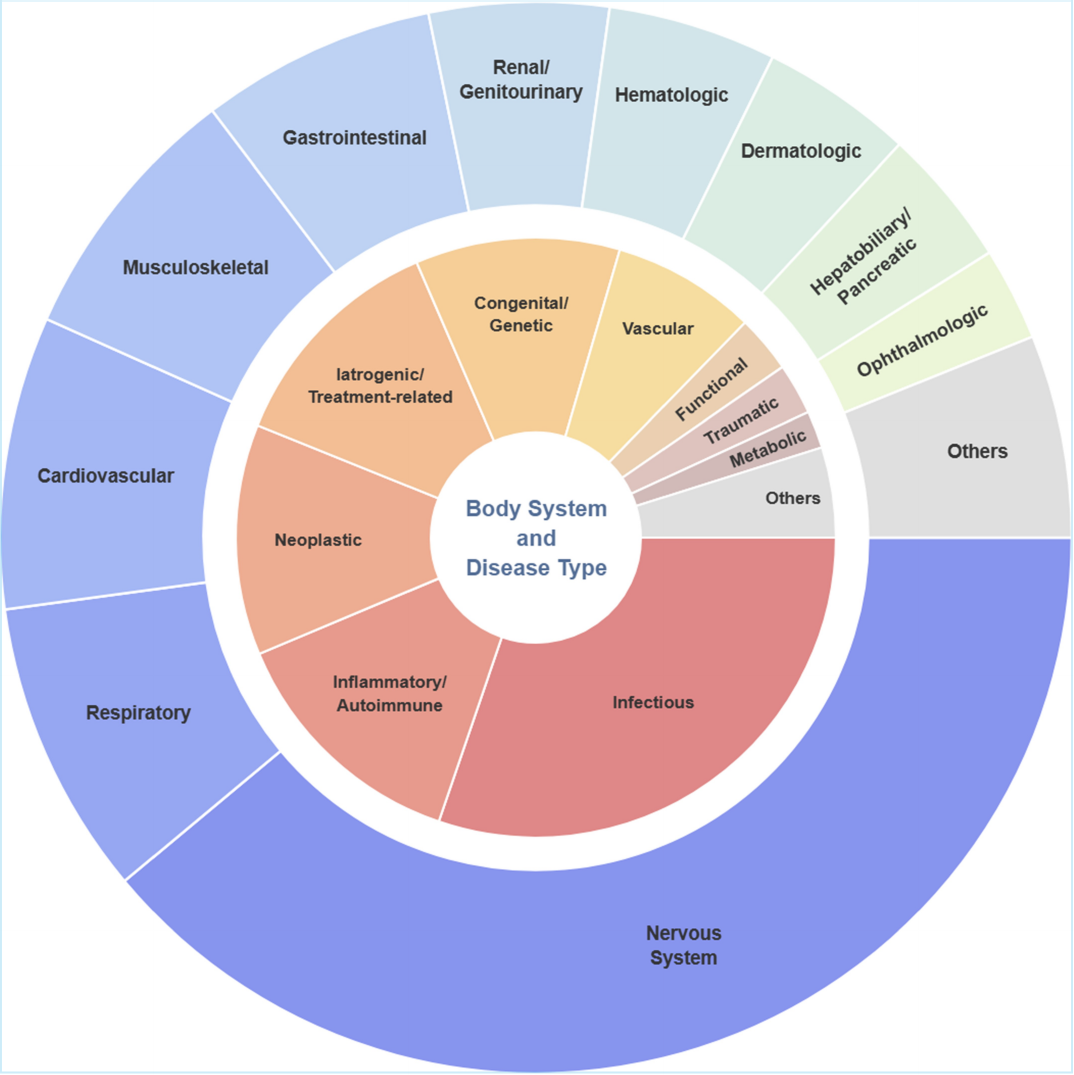}
    \caption{Dataset composition across primary body systems and disease types. The outer ring shows the distribution by primary body system, while the inner ring shows the distribution by disease type. The two rings represent independent category views and do not imply a hierarchical relationship.}
    \label{fig:appendix_data_comp}
\end{figure}

\begin{figure}[t]
    \centering
    \includegraphics[width=0.9\linewidth]{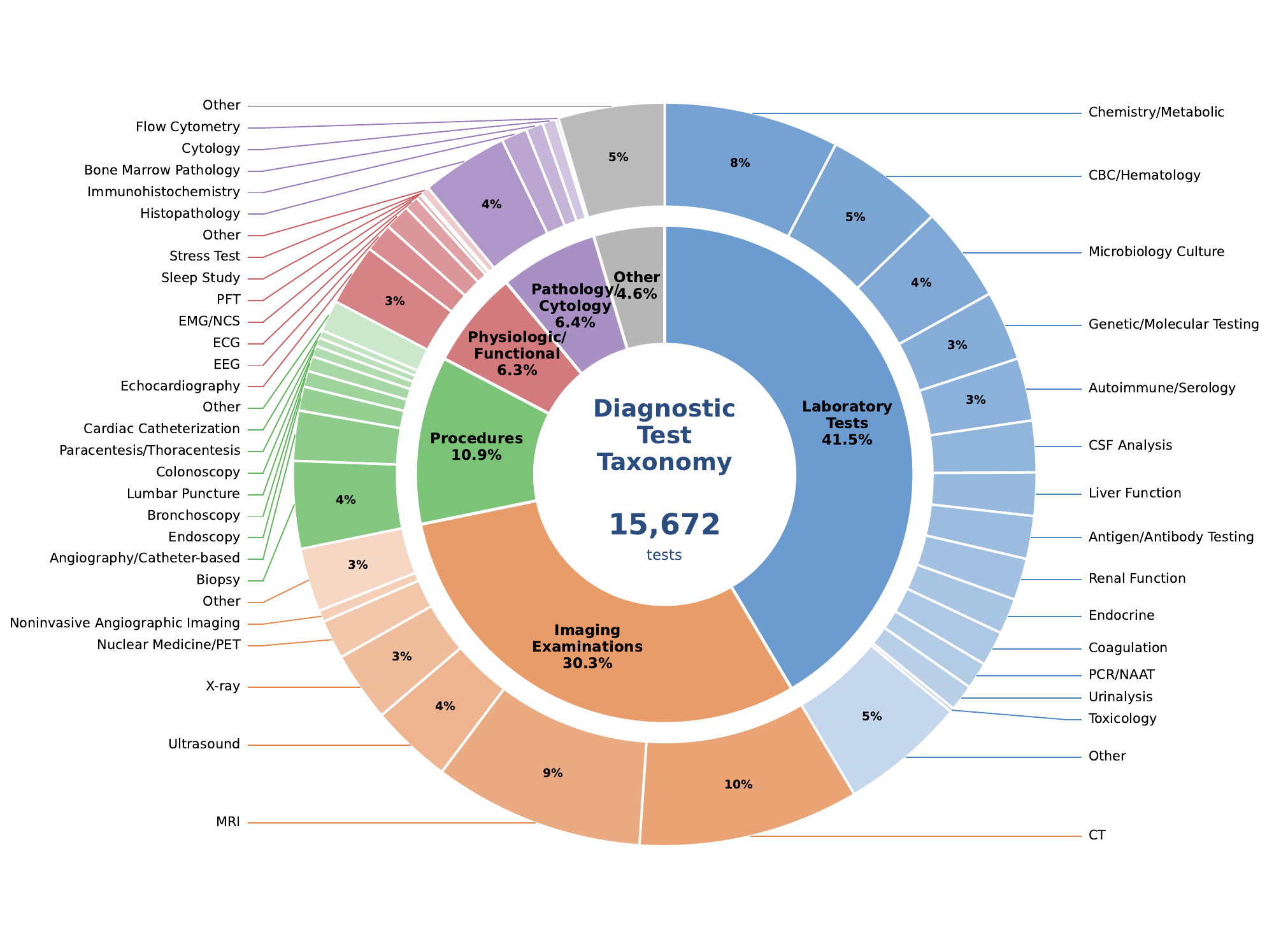}
    \caption{Distribution of diagnostic tests in the dataset. The inner ring shows major test categories, and the outer ring shows representative minor categories within each major category. The center number (15,672) denotes the total number of diagnostic tests in the dataset. Laboratory tests and imaging examinations account for the largest shares of the test space, followed by procedures/invasive diagnostic tests, physiologic/functional tests, and pathology/cytology.}
    \label{fig:appendix_test_dist}
\end{figure}

Figure~\ref{fig:appendix_data_comp} shows the distribution of the 2,896 source PMC case reports by primary body system and disease type. Figure~\ref{fig:appendix_test_dist} summarizes the distribution of diagnostic test categories across these cases.

\clearpage
\label{app:medaction300}
\paragraph{MedAction-300 Hard test dataset.} MedAction-300-Hard comprises 300 cases specifically selected for rare and uncommon diseases that are distinct from the other evaluation dataset, MedR-Bench. These cases are deliberately chosen because rare diseases pose a greater diagnostic challenge: they are underrepresented in training corpora, often mimic more common conditions, and typically require longer and more targeted test-ordering sequences to reach a definitive diagnosis. During construction, we used GPT-4o as a validator to ensure no data overlap between MedAction-32K, MedAction-300-Hard and MedR-Bench.
We evaluate our model on both the MedAction-300 Hard test set and the MedR-Bench test set. Compared with MedR-Bench, MedAction-300-Hard poses a substantially more challenging diagnostic setting. Cases in our benchmark require more diagnostic actions at each step, as shown by a higher average number of ordered tests per turn (1.92 vs. 1.57), and unfold over longer diagnostic trajectories, with a higher average number of turns per case (6.81 vs. 4.50). Together, these differences suggest that MedAction-300-Hard demands more extensive evidence acquisition and more sustained multi-turn clinical reasoning, making it a more rigorous benchmark for evaluating active diagnostic decision-making.

\section{Training and Inference Configs}
\label{app:training_config}

The max turns are set to 8. GPT-4o~\citep{hurst2024gpt} is used as the evaluator, to be consistent with original Med-R benchmark. We used a learning rate of 2e-4, 3 epochs, and a batch size of 32. Temperature was set to 0.6 and max output tokens set to 5500. All experiments were done on 4$\times$ NVIDIA RTX PRO 6000 Blackwell GPUs. 

For data generation, 3 rollout branches were generated initially per case,
1 branch launch points are then chosen to branch. DTC needs no threshold as it is a relative metric.
The RAC threshold is set to 3 as we found that the average RAC for correct samples in our teacher model Baichuan-M3 was 3.3. We also tried setting it to 4 which increased the data retention from 53\% to 78\% but yielded -3.5\% final performance on Med-R.

% ============================================================================
% [arXiv v2] Appendix D: MedAction-RL.
% INSERT: in main.tex, at the very end of the appendix, i.e. AFTER the
%         \section{Training and Inference Configs} section (after the
%         paragraph ending "...yielded -3.5\% final performance on Med-R.")
%         and BEFORE \end{document}.
%         i.e.  \clearpage
%               \input{appendix_medaction_rl}
%
% Figures used (place them in the project root or in figures/):
%   coldstart.pdf, scaling_law.pdf, data_scaling.pdf,
%   download.png, rl_test_acc_curve.pdf
% ============================================================================

\graphicspath{{./}{figures/}}

\section{MedAction-RL: Extending MedAction to Reinforcement Learning}
\label{app:rl}

\begin{tcolorbox}[colback=gray!6, colframe=gray!50, boxrule=0.5pt,
                  left=4pt, right=4pt, top=3pt, bottom=3pt]
\small\textbf{Note.} This appendix is a post-publication extension added in arXiv v2. It was not part of the COLM 2026 peer-reviewed version. The experiments here use a different agent interface and backbone from the main text (Appendix~\ref{app:rl_setup}), so their numbers are not directly comparable to Table~\ref{tab:results_compact}.
\end{tcolorbox}

This appendix details the reinforcement learning (RL) extension summarized in Section~\ref{sec:rl_extension}. It studies how MedAction's environments and DTC/RAC metrics can be used to post-train active diagnosis agents with SFT followed by RL. We address three questions: how to cold-start the policy (Appendix~\ref{app:rl_coldstart}), how performance scales with model capacity and the number of training environments (Appendix~\ref{app:rl_scaling}), and how to design rewards for the partially documented, noisy clinical environments derived from case reports (Appendix~\ref{app:rl_reward}, \ref{app:rl_reward_ablation}). Concurrent work also studies agentic training for clinical diagnosis, including conversational inquiry agents~\citep{lai2026doctorr}, multimodal medical agents~\citep{chen2026meissa}, and SFT+RL agents that ask, examine, and diagnose under noisy observations~\citep{gao2026medexagent}; here we focus on post-training design choices for test-ordering agents built on the MedAction pipeline.

% ----------------------------------------------------------------------------
\subsection{Setup}
\label{app:rl_setup}

\paragraph{Agent interface.}
Unlike the structured multi-section prompt used in the main text (Appendix~\ref{app:prompt_template}), the RL agent uses a tool-call interface (Appendix~\ref{app:rl_prompt}). At each turn the agent issues at most one \texttt{order\_test} tool call, writes brief reasoning, and outputs a concise current diagnosis and a \texttt{continue}/\texttt{done} status. The current diagnosis therefore plays the role of the top-ranked hypothesis $b_t^{(1)}$ in Eq.~\ref{eq:dtc}. The environment, oracle, and maximum of 8 turns follow Section~\ref{sec:data_gen}. Validation and evaluation use greedy decoding. Because the interface and decoding differ from the main text, results in this appendix should be compared only with each other.

\paragraph{Test-name matching.}
Agents generate free-form test names, which the environment resolves against each case's documented test menu $\mathcal{M}$ in four stages: (1) deterministic normalization and exact matching (capitalization, punctuation, spelling, and standard abbreviations); (2) expansion with curated clinical synonyms and naming variants; (3) for unresolved requests, retrieval of candidate tests from $\mathcal{M}$ with MedCPT~\citep{jin2023medcpt}, whose similarity score is used only for candidate retrieval and low-confidence rejection rather than as evidence of equivalence; and (4) a binary equivalence judgment between the request and the retrieved candidates by Qwen3.5-9B. Requests that remain unmatched return \texttt{UNAVAILABLE}.

\paragraph{Cold-start SFT.}
The cold-start data are the 32,681 MedAction training instances (Section~\ref{sec:data_gen}), generated with multiple teachers and tree-structured sampling and filtered with DTC and RAC. We compare two SFT formats. In \textit{turn-level} SFT, each diagnostic turn is an independent training instance: the input is the accumulated context $h_t=(o_1,a_1,b_1,r_1,\dots,a_{t-1},b_{t-1},r_{t-1})$ and the target is the teacher response $(a_t,b_t)$, i.e., the ordered test and updated belief, or the committed diagnosis $b_T$ at the final turn; the loss is normalized per turn. In \textit{trajectory-level} SFT, the full teacher trajectory is serialized into one sequence, and the loss is applied only to model-generated tokens.

\paragraph{RL training and evaluation.}
Unless stated otherwise, the policy is Qwen3.5-27B (dense, thinking enabled)~\citep{qwen35}, cold-started with turn-level SFT and then trained with GRPO~\citep{shao2024deepseekmath} using 8 rollouts per case on the 2,896 MedAction environments. We evaluate on MedR-Bench and MedAction-300-Hard with the same metrics as the main text (test-recommendation precision, recall, F1, and diagnostic accuracy judged by GPT-4o, averaged over 3 runs), and perform case-level overlap checks between training and evaluation data. Note that precision, recall, and F1 measure overlap with the tests documented in each case report, not clinical appropriateness. Full hyperparameters are listed in Appendix~\ref{app:rl_impl}. Throughout this appendix, we refer to II-Medical-8B~\citep{2025II-Medical-8B}, the Qwen3-8B-based medical model used in the main text, as Qwen3-8B Medical.

% ----------------------------------------------------------------------------
\subsection{Reward Design}
\label{app:rl_reward}

\paragraph{Diagnosis reward.}
An LLM-based verifier checks whether the final diagnosis $b_T$ matches the ground truth $s^*$:
\begin{equation}
\label{eq:rl_diag}
R_{\mathrm{diag}}(\tau) = \mathbb{1}\!\left[\mathrm{Verify}(b_T, s^*) = \textsc{correct}\right].
\end{equation}

\paragraph{DTC-gated test reward.}
A natural dense reward for test ordering is the Jaccard overlap between the set of ordered tests $\mathcal{A}_\tau=\bigcup_{t=1}^{T} a_t$ and the documented menu $\mathcal{M}$, i.e., $|\mathcal{A}_\tau\cap\mathcal{M}|/|\mathcal{A}_\tau\cup\mathcal{M}|$. However, $\mathcal{M}$ is not a minimal, clinically indicated workup: it can contain redundant or management-oriented tests, while reasonable tests absent from the case report return \texttt{UNAVAILABLE}. Maximizing overlap with $\mathcal{M}$ therefore rewards coverage of the case inventory rather than tests that actually help the agent reach a better diagnosis.

We instead credit a test according to its effect on the agent's own diagnostic trajectory, reusing the DTC metric (Eq.~\ref{eq:dtc}). Let $\delta_t = \mathrm{DTC}_t - \mathrm{DTC}_{t+1}$ be the change in PrimeKG distance between the agent's hypothesis and $s^*$ after the result of $a_t$ is returned. We keep only tests that preserve or improve diagnostic proximity, $\widetilde{\mathcal{A}}_\tau = \{a_t \in \mathcal{A}_\tau : \delta_t \ge 0\}$, and define
\begin{equation}
\label{eq:rl_test}
R_{\mathrm{test}}(\tau) = \frac{|\widetilde{\mathcal{A}}_\tau \cap \mathcal{M}|}{|\mathcal{A}_\tau \cup \mathcal{M}|}.
\end{equation}
We use the non-regression criterion $\delta_t \ge 0$ rather than $\delta_t > 0$ so as not to penalize trajectories that are already correct or near-correct early, where further graph-distance improvement is small or impossible. Rewarding only strict improvement can also encourage the agent to keep switching its belief in search of additional progress instead of stopping decisively.

\paragraph{Total reward.}
During training, the policy sometimes re-ordered semantically equivalent tests under different names, so we add an LLM-judged repetition penalty $P_{\mathrm{rep}}(\tau)$ that measures semantic redundancy among ordered tests, as well as a format reward $R_{\mathrm{format}}(\tau)$ for parsable outputs. The trajectory reward is
\begin{equation}
\label{eq:rl_total}
R(\tau) = R_{\mathrm{diag}}(\tau) + \min\!\big(R_{\mathrm{test}}(\tau),\, \alpha\big) - \lambda_{\mathrm{rep}} P_{\mathrm{rep}}(\tau) + \lambda_{\mathrm{format}} R_{\mathrm{format}}(\tau),
\end{equation}
with $\alpha = 0.4$, $\lambda_{\mathrm{rep}} = 0.02$, and $\lambda_{\mathrm{format}} = 0.1$. Clamping the test reward from above keeps the auxiliary signal from dominating the diagnosis reward; Appendix~\ref{app:rl_impl} compares it with multiplicative scaling. An undocumented test receives no credit in Eq.~\ref{eq:rl_test} but no explicit negative reward, and does not prevent the agent from obtaining the full diagnosis reward.

% ----------------------------------------------------------------------------
\subsection{Main Results}
\label{app:rl_main}

\begin{table}[h]
\centering
\small
\setlength{\tabcolsep}{4pt}
\resizebox{\textwidth}{!}{%
\begin{tabular}{l cccc cccc}
\toprule
& \multicolumn{4}{c}{\textbf{MedR-Bench}} & \multicolumn{4}{c}{\textbf{MedAction-300-Hard}} \\
\cmidrule(lr){2-5} \cmidrule(lr){6-9}
\textbf{Model} & Prec & Rec & F1 & Diag Acc & Prec & Rec & F1 & Diag Acc \\
\midrule
\multicolumn{9}{l}{\textit{Closed-Source Models}} \\
OpenAI-o3-mini        & \underline{0.36} & 0.42 & 0.39 & \underline{0.71} & \textbf{0.54} & 0.47 & \textbf{0.50} & \underline{0.67} \\
Gemini 2.5 Flash-Lite & \underline{0.36} & \underline{0.44} & \underline{0.40} & 0.68 & \underline{0.47} & \underline{0.48} & 0.47 & 0.63 \\
GPT-5.4               & \textbf{0.47} & \textbf{0.56} & \textbf{0.51} & \textbf{0.81} & 0.42 & \textbf{0.60} & \underline{0.49} & \textbf{0.82} \\
\midrule
\multicolumn{9}{l}{\textit{Open-Source Models}} \\
Baichuan-M1 (14B)     & 0.42 & 0.39 & 0.40 & 0.61 & 0.47 & 0.41 & 0.44 & 0.56 \\
Baichuan-M3 (235B)    & 0.52 & \textbf{0.52} & \underline{0.52} & \underline{0.69} & 0.50 & \textbf{0.54} & \underline{0.52} & \underline{0.66} \\
MedGemma (27B)        & 0.39 & 0.46 & 0.42 & 0.65 & \underline{0.52} & \underline{0.52} & \underline{0.52} & 0.63 \\
Qwen-QwQ (32B)        & 0.35 & 0.49 & 0.41 & 0.66 & 0.51 & 0.41 & 0.45 & 0.58 \\
Qwen3-8B Medical         & \underline{0.59} & 0.36 & 0.45 & 0.51 & 0.45 & 0.45 & 0.45 & 0.46 \\
Qwen3.5-27B           & 0.49 & 0.36 & 0.42 & 0.46 & 0.45 & 0.41 & 0.43 & 0.45 \\
\midrule
MedAction-RL (27B)    & \textbf{0.65} & \underline{0.51} & \textbf{0.57} & \textbf{0.77} & \textbf{0.62} & 0.48 & \textbf{0.54} & \textbf{0.73} \\
\bottomrule
\end{tabular}%
}
\caption{RL results on MedR-Bench and MedAction-300-Hard (averaged over 3 runs). Rows other than Qwen3.5-27B and MedAction-RL are reproduced from Table~\ref{tab:results_compact}. Bold and underline denote the best and second-best results within each model group.}
\label{tab:rl_main}
\end{table}

As shown in Table~\ref{tab:rl_main}, MedAction-RL (27B) improves diagnostic accuracy over its Qwen3.5-27B base from 0.46 to 0.77 on MedR-Bench and from 0.45 to 0.73 on MedAction-300-Hard (mean $\pm$ std over 3 runs: $0.770 \pm 0.024$ and $0.730 \pm 0.027$, versus $0.460 \pm 0.025$ and $0.450 \pm 0.025$ for the base). It obtains the highest diagnostic accuracy among the open-source models listed, while GPT-5.4 remains the strongest model overall. MedAction-RL also attains the highest test-recommendation precision among these models, with lower recall than some baselines. This is consistent with an agent that orders a small, targeted set of tests under the one-test-per-turn interface rather than broad panels. The baselines also show that better test overlap alone does not guarantee better diagnosis (e.g., GPT-5.4 has moderate overlap scores but the best diagnostic accuracy): active diagnosis requires both selecting useful tests and correctly interpreting their results, which motivates crediting only tests that do not regress the diagnosis (Eq.~\ref{eq:rl_test}).

% ----------------------------------------------------------------------------
\subsection{Cold-Starting the Policy}
\label{app:rl_coldstart}

\begin{figure}[h]
\centering
\includegraphics[width=0.85\linewidth]{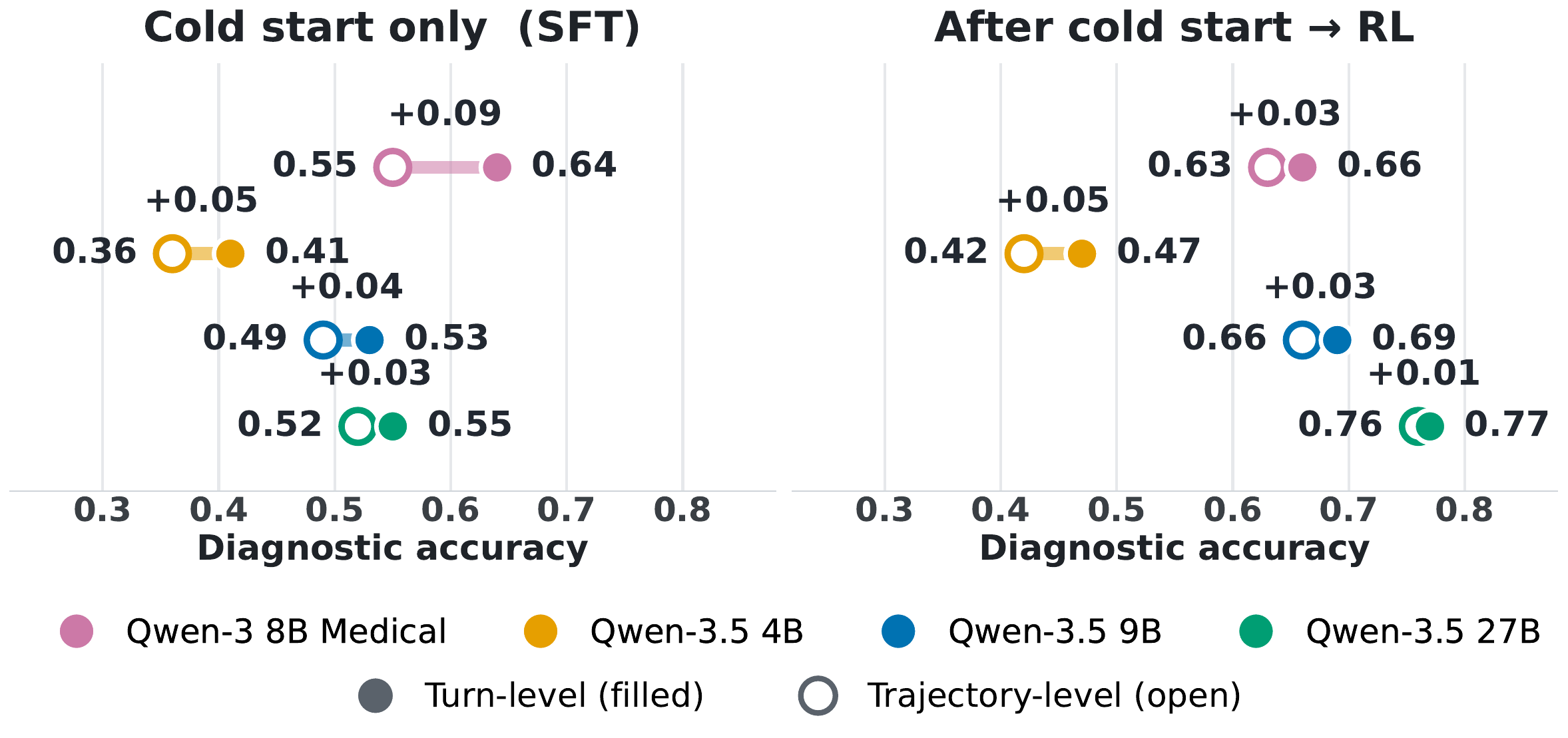}
\caption{MedR-Bench diagnostic accuracy after cold-start SFT (left) and after subsequent RL (right) for four backbones. Filled markers: turn-level SFT on DTC/RAC-filtered data; open markers: trajectory-level SFT on outcome-filtered data.}
\label{fig:rl_coldstart}
\end{figure}

\begin{table}[h]
\centering
\small
\begin{tabular}{ll cc cc}
\toprule
& & \multicolumn{2}{c}{\textbf{Qwen3.5-9B} (base 0.44)} & \multicolumn{2}{c}{\textbf{Qwen3.5-27B} (base 0.46)} \\
\cmidrule(lr){3-4} \cmidrule(lr){5-6}
\textbf{Filtering} & \textbf{Format} & SFT & SFT+RL & SFT & SFT+RL \\
\midrule
Outcome-only & Trajectory & 0.49 & 0.66 & 0.52 & 0.76 \\
Outcome-only & Turn       & 0.47 & 0.64 & 0.53 & 0.75 \\
DTC/RAC      & Turn       & \textbf{0.53} & \textbf{0.69} & \textbf{0.55} & \textbf{0.77} \\
\bottomrule
\end{tabular}
\caption{Controlled cold-start ablation separating the effect of SFT format from that of data filtering (MedR-Bench diagnostic accuracy).}
\label{tab:rl_coldstart_ablation}
\end{table}

Figure~\ref{fig:rl_coldstart} compares turn-level SFT on DTC/RAC-filtered data with trajectory-level SFT on outcome-filtered data (trajectories that reach the correct diagnosis). Across all four backbones, the former gives a stronger initialization, and the advantage largely persists after RL. To separate the two factors, Table~\ref{tab:rl_coldstart_ablation} adds a control that uses turn-level formatting with outcome-only filtering. Turn-level formatting alone does not outperform trajectory-level SFT; the gain appears only when turn-level training is combined with DTC/RAC filtering. Turn-level instances are what make this fine-grained curation possible: they allow us to keep convergent prefixes and individually grounded turns from otherwise imperfect trajectories, rather than accepting or rejecting whole trajectories by their final outcome.

\begin{tcolorbox}[colback=blue!5, colframe=blue!30, boxrule=0.5pt,
                  left=4pt, right=4pt, top=3pt, bottom=3pt]
\small\textbf{Takeaway:} The cold-start benefit comes mainly from fine-grained DTC/RAC curation, which turn-level training enables; turn-level formatting alone is not sufficient.
\end{tcolorbox}

% ----------------------------------------------------------------------------
\subsection{Scaling Model Size and Training Environments}
\label{app:rl_scaling}

\begin{figure}[h]
\centering
\includegraphics[width=0.8\linewidth,trim={0 5pt 0 5pt},clip]{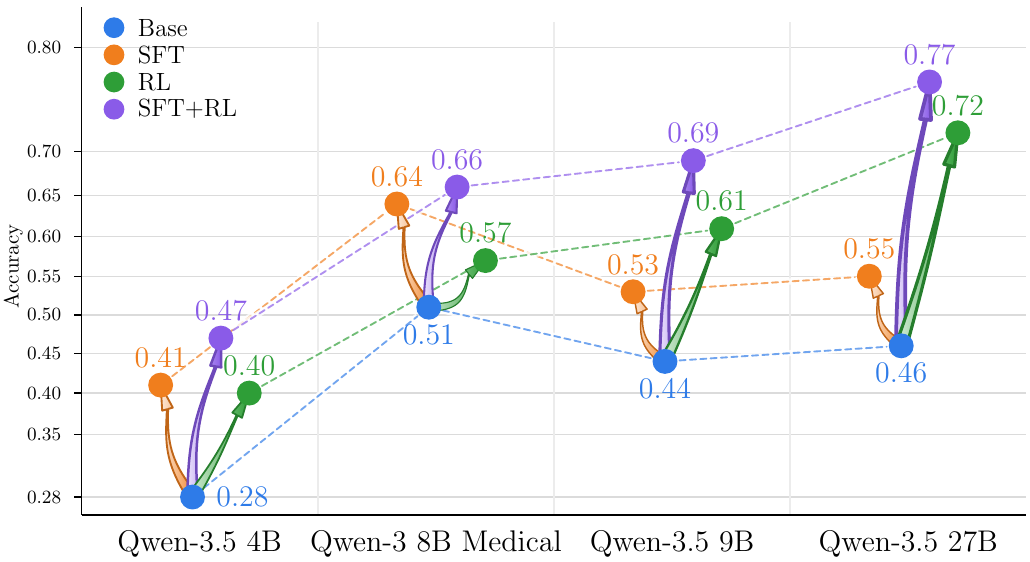}\\[2pt]
\includegraphics[width=0.8\linewidth,trim={0 5pt 0 5pt},clip]{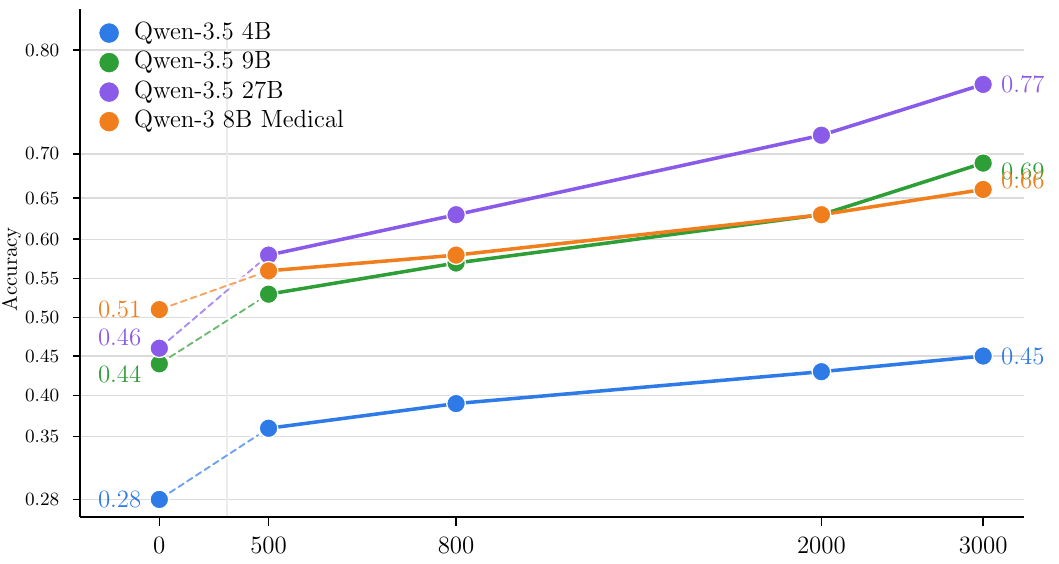}
\caption{Top: MedR-Bench diagnostic accuracy of Base, SFT, RL, and SFT+RL for four backbones. Bottom: diagnostic accuracy as a function of the number of training cases (0 corresponds to the base model).}
\label{fig:rl_scaling}
\end{figure}

\begin{table}[h]
\centering
\small
\begin{tabular}{l cccc}
\toprule
\textbf{Backbone} & Base & SFT & RL & SFT+RL \\
\midrule
Qwen3.5-4B          & 0.28 & 0.41 & 0.40 & 0.47 \\
Qwen3.5-9B          & 0.44 & 0.53 & 0.61 & 0.69 \\
Qwen3.5-27B         & 0.46 & 0.55 & 0.72 & 0.77 \\
\midrule
Qwen3-8B Medical    & 0.51 & 0.64 & 0.57 & 0.66 \\
MedGemma-4B         & 0.24 & 0.39 & 0.43 & 0.46 \\
MedGemma-27B        & 0.65 & 0.68 & 0.71 & 0.74 \\
\bottomrule
\end{tabular}
\caption{MedR-Bench diagnostic accuracy across backbones and training configurations. ``RL'' denotes RL from the base model without SFT.}
\label{tab:rl_scaling}
\end{table}

Figure~\ref{fig:rl_scaling} and Table~\ref{tab:rl_scaling} summarize how performance changes with model capacity and the number of training environments.

\textbf{SFT gives a roughly size-invariant gain; RL benefits more from capacity.}
Within the Qwen3.5 family, SFT adds a similar gain (about $+0.09$ to $+0.13$) at every size, whereas the gain from RL grows with model size, both on its own (from $+0.12$ at 4B to $+0.26$ at 27B) and on top of SFT (from $+0.06$ at 4B to $+0.22$ at 27B). Qwen3.5-9B and Qwen3.5-27B start from similar base accuracy (0.44 vs.\ 0.46), yet the larger model gains much more from RL: larger general-domain backbones respond more strongly to reinforcement.

\textbf{Medical backbones start stronger but gain less.}
Qwen3-8B Medical has a higher base accuracy than Qwen3.5-27B, but gains less from RL and is overtaken by Qwen3.5-9B after SFT+RL. MedGemma-27B shows the same pattern: it starts far above Qwen3.5-27B (0.65 vs.\ 0.46) yet improves by only $+0.09$ after SFT+RL, compared with $+0.31$. One interpretation is that domain-specific training already concentrates the model's distribution on medicine, leaving less headroom for RL. We note that these observations cover only a few sizes from two model families and should not be read as a general scaling law.

\textbf{More training environments help.}
Increasing the number of training environments improves accuracy for all backbones (Figure~\ref{fig:rl_scaling}, bottom), with larger gains for the stronger general-domain models.

\begin{tcolorbox}[colback=blue!5, colframe=blue!30, boxrule=0.5pt,
                  left=4pt, right=4pt, top=3pt, bottom=3pt]
\small\textbf{Takeaway:} Performance improves with model capacity and the number of environments, but the gains from post-training depend on the backbone: RL helps larger general-domain models most, while medical backbones start higher and gain less.
\end{tcolorbox}

% ----------------------------------------------------------------------------
\subsection{Reward Ablation}
\label{app:rl_reward_ablation}

\begin{table}[h]
\centering
\small
\begin{tabular}{lcc}
\toprule
\textbf{Test reward} & \textbf{MedR-Bench} & \textbf{MedAction-300-Hard} \\
\midrule
None                         & 0.69 & 0.64 \\
Jaccard overlap              & 0.75 & \textbf{0.73} \\
DTC-gated (Eq.~\ref{eq:rl_test}) & \textbf{0.77} & \textbf{0.73} \\
\bottomrule
\end{tabular}
\caption{Diagnostic accuracy of Qwen3.5-27B (SFT+RL) trained with different test rewards.}
\label{tab:rl_reward_ablation}
\end{table}

\begin{figure}[h]
\centering
\includegraphics[width=0.9\linewidth]{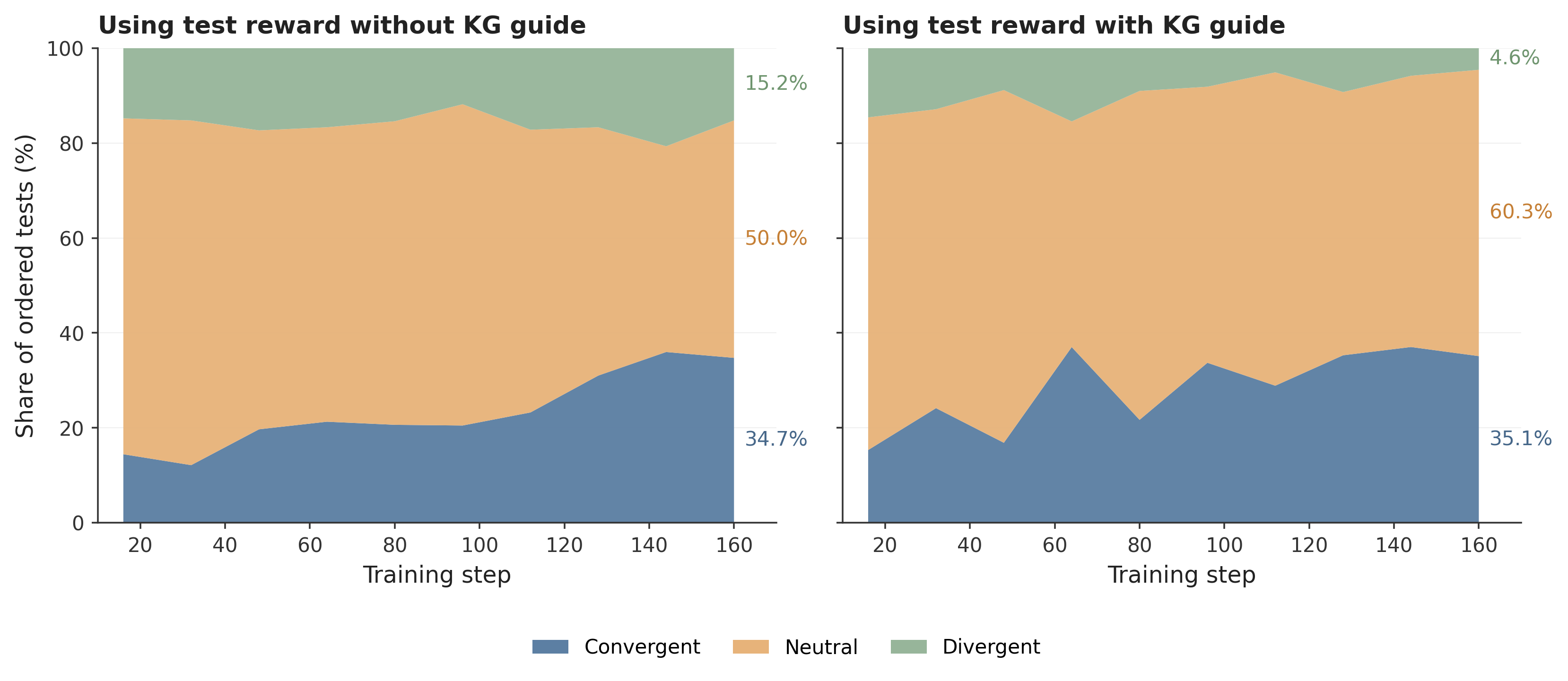}
\caption{Share of ordered tests followed by convergent, neutral, or divergent belief updates (DTC decreases, stays the same, or increases) over RL training. Left: Jaccard test reward. Right: DTC-gated test reward.}
\label{fig:rl_reward_composition}
\end{figure}

Table~\ref{tab:rl_reward_ablation} compares training without a test reward, with the Jaccard overlap reward, and with the DTC-gated reward. Adding a test reward is clearly beneficial: with the diagnosis reward alone, accuracy is 0.69 and 0.64. The DTC-gated reward improves over Jaccard on MedR-Bench and ties it on MedAction-300-Hard. Its main effect appears in trajectory behavior (Figure~\ref{fig:rl_reward_composition}). Both rewards lead to a similar share of convergent tests by the end of training (34.7\% with Jaccard vs.\ 35.1\% with ours), but the DTC-gated reward reduces the share of tests followed by divergent belief updates from 15.2\% to 4.6\%. In other words, at comparable accuracy, the agent reaches its diagnosis with far fewer belief-regressing test orders. The large neutral share under both rewards also shows that many correctly ordered tests still do not change the agent's belief, pointing to evidence interpretation as a remaining bottleneck.

\begin{tcolorbox}[colback=blue!5, colframe=blue!30, boxrule=0.5pt,
                  left=4pt, right=4pt, top=3pt, bottom=3pt]
\small\textbf{Takeaway:} A dense test reward is necessary, and gating it with DTC reduces belief-regressing test orders without sacrificing accuracy.
\end{tcolorbox}

% ----------------------------------------------------------------------------
\subsection{Test Cost}
\label{app:rl_cost}

\begin{table}[h]
\centering
\small
\resizebox{\linewidth}{!}{%
\begin{tabular}{l c ccccc c}
\toprule
& & \multicolumn{5}{c}{\textbf{Mean cumulative cost by turn (USD)}} & \\
\cmidrule(lr){3-7}
\textbf{Model} & \textbf{Avg.\ turns} & 1 & 2 & 3 & 4 & $\ge$5 & \textbf{Avg.\ cost/case} \\
\midrule
Qwen3.5-27B (base)   & 3.02 & 352 & 631 & 922 & 1,142 & \textbf{1,221} & \textbf{840.58} \\
OpenAI-o3-mini       & 4.12 & 305 & 635 & 951 & 1,283 & 1,521 & 1,162.40 \\
MedAction-RL (27B)   & 4.14 & \textbf{264} & \textbf{549} & \textbf{823} & \textbf{1,110} & 1,316 & 1,010.63 \\
\bottomrule
\end{tabular}%
}
\caption{Estimated test cost on 100 MedAction-300-Hard cases, priced with CMS fee schedules~\citep{cms2025feeschedule}. Diagnostic accuracy (GPT-4o judged) on these cases is 0.46, 0.62, and 0.70, respectively.}
\label{tab:rl_cost}
\end{table}

We estimate the cost of ordered tests using the U.S.\ Centers for Medicare \& Medicaid Services (CMS) fee schedules~\citep{cms2025feeschedule}. These estimates are directional, since actual costs vary across institutions, countries, and insurance plans. As shown in Table~\ref{tab:rl_cost}, MedAction-RL has a lower cumulative cost than the base model through turn 4 and exceeds it only from turn 5 onward, because it runs longer workups (4.14 vs.\ 3.02 turns on average). Compared with o3-mini, which takes a similar number of turns, MedAction-RL incurs lower cost at every turn and about 13\% lower cost per case, while achieving higher diagnostic accuracy. The gains in diagnostic accuracy therefore do not come from ordering more expensive tests.

% ----------------------------------------------------------------------------
\subsection{Implementation Details and Additional Findings}
\label{app:rl_impl}

\paragraph{Hyperparameters.}
Table~\ref{tab:rl_hparams} lists the RL configuration for the main Qwen3.5-27B run; hyperparameters are kept fixed across runs. Training runs for 3 epochs with early stopping on a 10\% validation split. Rollouts are generated with vLLM~\citep{kwon2023efficient}. Our compute pool consists of 4 nodes with 8 NVIDIA H200 GPUs each; each individual experiment runs on a single node (8 H200 GPUs) using FSDP with parameter offloading enabled and optimizer offloading disabled. The full study used over 15,000 GPU hours.

\begin{table}[h]
\centering
\small
\begin{tabular}{@{}l>{\raggedright\arraybackslash}p{0.58\linewidth}@{}}
\toprule
\textbf{Setting} & \textbf{Value} \\
\midrule
Algorithm & GRPO with 8 rollouts per case \\
Training data & 2,896 environments, 3 epochs \\
Batch size (train / mini / micro) & 24 / 12 / 1 \\
Optimizer & AdamW, lr $1\times10^{-6}$, $\beta=(0.9, 0.999)$, weight decay 0.01 \\
LR schedule & Constant, no warmup \\
Gradient clipping & 1.0 \\
Policy clipping & PPO clip ratio 0.2, dual-clip constant $c=3.0$ \\
Advantage / loss & Group-std-normalized advantages; token-mean loss aggregation \\
KL loss & Coefficient 0.001, low-variance KL estimator \\
Entropy coefficient & 0.001 \\
Precision & bf16 compute, fp32 master weights; gradient checkpointing \\
Max turns & 8 \\
Max prompt / response length & 2,048 / 8,192 tokens \\
Tool response length & Truncated to 256 tokens (middle truncation) \\
Rollout temperature & 1.0 (training); greedy, temperature 0 (validation) \\
Agent-loop workers & 8 \\
Reward weights & $\alpha=0.4$, $\lambda_{\mathrm{rep}}=0.02$, $\lambda_{\mathrm{format}}=0.1$ \\
Evaluator & GPT-4o~\citep{hurst2024gpt} \\
\bottomrule
\end{tabular}
\caption{RL hyperparameters.}
\label{tab:rl_hparams}
\end{table}

\paragraph{Test-reward shaping.}
We compared two ways to limit the weight of the test reward: multiplicative scaling $\alpha \cdot R_{\mathrm{test}}$ with $\alpha \in \{0.1, 0.5, 1\}$, and an upper clamp $\min(R_{\mathrm{test}}, \alpha)$ with $\alpha \in \{0.2, 0.4\}$. On MedR-Bench, clamping at $\alpha=0.4$ performed best, and we use it in all experiments.

\paragraph{Entropy regularization prevents test-ordering collapse.}
Without entropy regularization, the policy tends to collapse to a small set of generic test-ordering templates, repeatedly ordering the same high-frequency panels regardless of the patient's presentation. A small entropy coefficient (0.001) preserves diversity in diagnostic actions and prevents this collapse.

\paragraph{Similarity-based dense rewards are easily hacked.}
A natural way to densify the reward is to use embedding similarity (e.g., MedCPT~\citep{jin2023medcpt}) to give partial credit for ordered tests that resemble documented tests, or for diagnoses that resemble but do not match the ground truth. In our experiments this consistently led to reward hacking: the policy learned to order tests that are semantically similar but diagnostically distinct, inflating the reward without improving diagnosis, or quickly converged on diagnoses in the correct organ system that were nevertheless wrong.

\begin{figure}[h]
\centering
\includegraphics[width=0.75\linewidth]{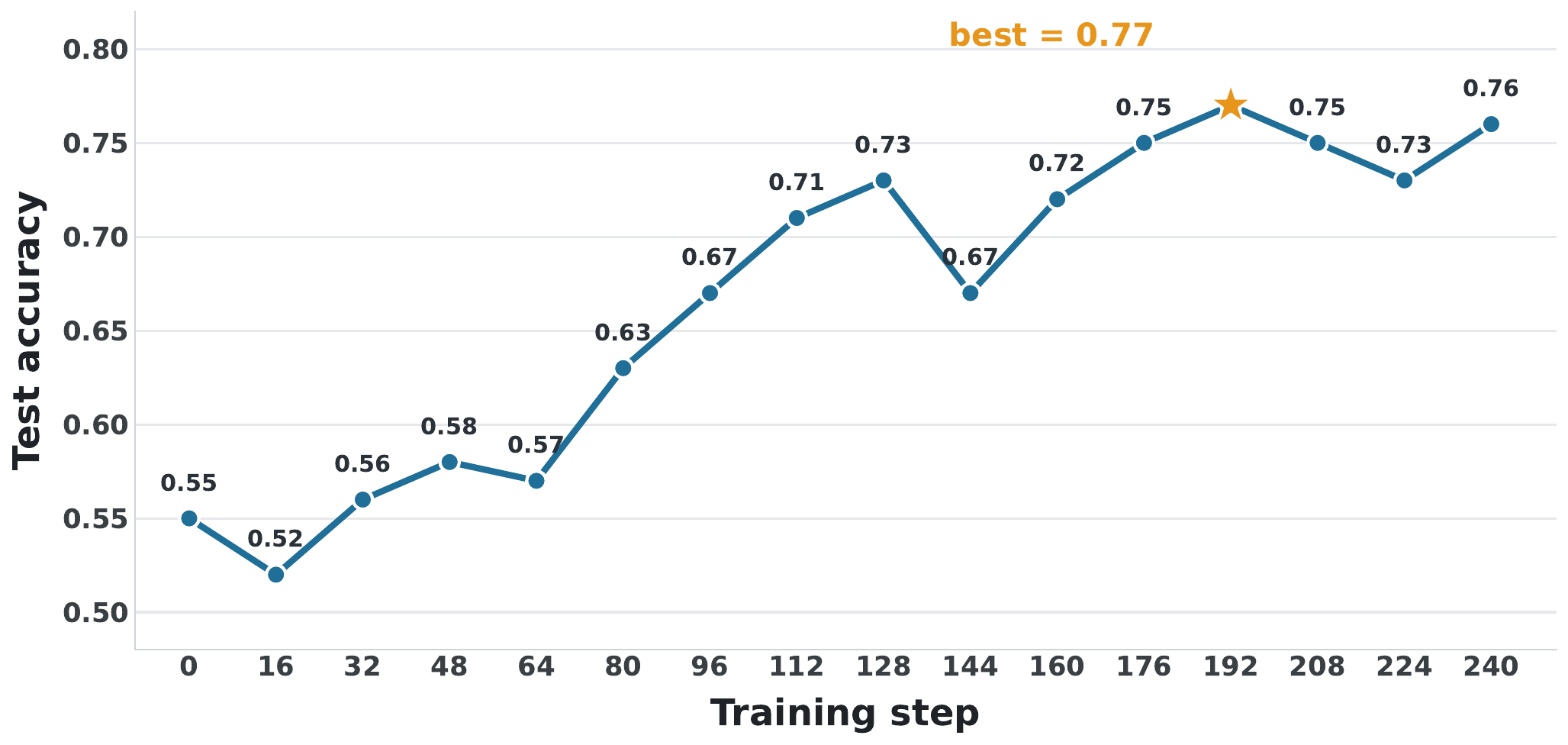}
\caption{MedR-Bench diagnostic accuracy over RL training steps (Qwen3.5-27B, SFT+RL).}
\label{fig:rl_curve}
\end{figure}

\paragraph{Training dynamics.}
Figure~\ref{fig:rl_curve} shows MedR-Bench diagnostic accuracy over the course of RL training for the main run, which rises from 0.55 after cold-start SFT to a best of 0.77.

% ----------------------------------------------------------------------------
\subsection{Limitations of the RL Extension}
\label{app:rl_limitations}

The environments used here are derived from case reports, which selectively document unusual or diagnosis-relevant findings and do not define a complete clinical action space. The simulator therefore cannot return patient-specific results for many clinically appropriate routine tests; such tests return \texttt{UNAVAILABLE}, receive no credit in Eq.~\ref{eq:rl_test}, and consume an interaction. The DTC-gated, clamped test reward reduces but does not eliminate this dependence on the documented test inventory, and the overlap-based precision, recall, and F1 should be read accordingly. All results in this appendix reflect performance in a case-report-derived simulator rather than in real clinical decision-making; evaluation on EHR-derived or clinician-authored environments, and within real clinical workflows, would be needed before broader claims about active diagnostic agents. MedAction-RL is a research system and is not intended for autonomous clinical use.

\end{document}